\documentclass[journal]{IEEEtran}

\usepackage{graphicx}
\DeclareGraphicsExtensions{.pdf,.jpg,.jpeg,.png}
\usepackage{amsmath}
\usepackage{amssymb}
\usepackage{array}
\usepackage{multirow}
\usepackage{ragged2e}
\usepackage{xcolor}
\usepackage{fvextra}
\usepackage[linesnumbered,ruled,vlined]{algorithm2e}
\usepackage{float}
\usepackage{morefloats}
\usepackage{stfloats}
\usepackage{placeins}
\usepackage{siunitx}
\usepackage{setspace}
\usepackage{tikz}
\usepackage{pgfplots}
\usetikzlibrary{arrows.meta,positioning,fit,calc,matrix,shapes.geometric,intersections}
\usepgfplotslibrary{groupplots}
\pgfplotsset{compat=1.18}

\usepackage[hidelinks]{hyperref}

\usepackage[comma,numbers]{natbib}

\title{A System for Fast, Resilient, and Adaptable Loco-Manipulation Behaviors on Humanoid Robots}

\author{Duncan Calvert$^{1,2}$,
    Luigi Penco$^{2}$,
    Dexton Anderson$^{2}$,
    Tomasz Bialek$^{2}$ \\
    Arghya Chatterjee$^{1}$,
    Beomyeong Park$^{2}$,
    and Robert Griffin$^{1,2}$
\thanks{This work was supported in part by ONR Grants N000-14-19-1-2023 (Fast Behaviors) and N00014-22-1-2593 (SquadBot v2), ONR Contract N00014-25-C2410 (SquadBot v3), and Collaborative Agreement W911NF2220201 (Breaching).}%
\thanks{$^{1}$Duncan Calvert, Arghya Chatterjee, and Robert Griffin are with the Florida Institute for Human and Machine Cognition, 40 S Alcaniz St, Pensacola, FL 32502, USA.}%
\thanks{$^{2}$Luigi Penco, Dexton Anderson, Tomasz Bialek, Beomyeong Park, and Robert Griffin are with the University of West Florida, 11000 University Pkwy, Pensacola, FL 32514, USA.}%
\thanks{Email: \{dcalvert, lpenco, danderson, tbialek, achatterjee, bpratt, rgriffin\}@ihmc.org.}%
}

\begin{document}

\maketitle

\phantomsection\label{ch:building}
\phantomsection\label{ch:tutorial}
\phantomsection\label{ch:metrics}
\phantomsection\label{sec:evaluation}
\phantomsection\label{sec:drc_era}
\phantomsection\label{sec:simple_behavior_json_file}
\phantomsection\label{sec:fallback_node}
\phantomsection\label{sec:authoring_a_frame_based_arm_action}

\begin{abstract}
There is tremendous value in humanoid robots taking on physically demanding, hazardous, and repetitive work in spaces built for humans.
However, a useful robot for these spaces must coordinate locomotion, whole-body motion, perception, contact, and operator supervision.
We present a robot-local, runtime-editable behavior authoring and runtime system that addresses these challenges.
We argue that behavior architecture can be a primary enabler of capability, speed, and reliability, and that runtime editability enables fast behavior creation, adaptation, extension, and combination.

Our behavior architecture combines object-centric Affordance Templates, a tree structure that provides organization and logic, and runtime-editable perception through a behavior scene and primitive scene actions.
Our operator interface remains continuously synchronized to the robot for runtime authoring, monitoring, and repair.
Action primitives execute through a whole-body controller that supports concurrent body motions and walking.

Demonstrations of our system cover six task variants on Unitree H1-2 and Alex.
We execute a push door traversal in 34 seconds and sort six balls by color in 45 seconds under human disturbance.
Timed authoring sessions show scratch creation of new loco-manipulation behaviors and adaptation of existing ones in hours.
Comparison against the literature finds our approach to be competitive with recent learned systems.

\end{abstract}


\section{Introduction}
\label{ch:introduction2}
\begin{figure}[t]
    \centering
    \includegraphics[width=0.95\columnwidth]{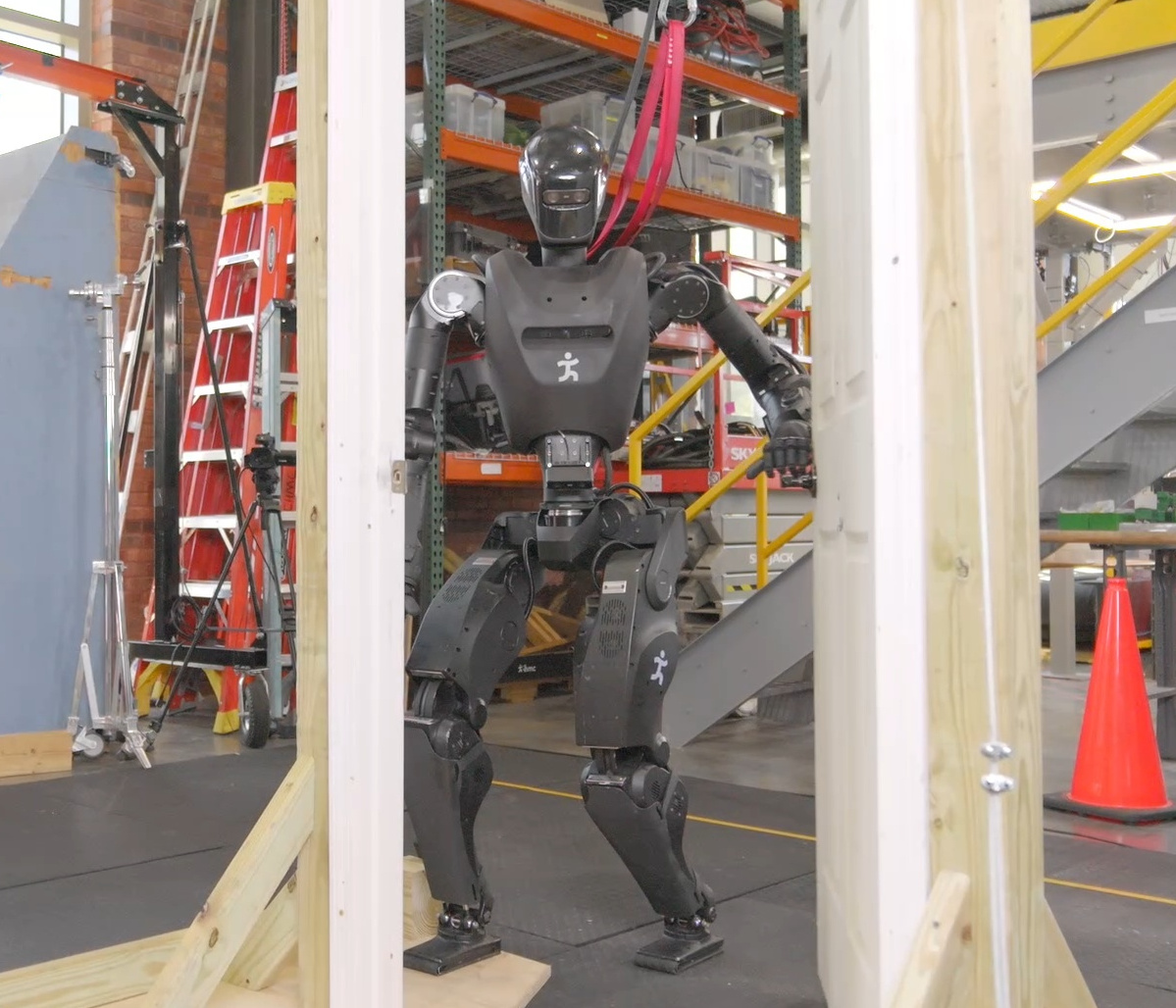}
    \caption{IHMC's fully electric Alex humanoid robot traversing a right pull door.
    Alex is the primary platform in the evaluations presented here. Supplementary video material accompanies this submission.}
    \label{fig:20260309_AlexRightPullDoorProfessionalStill}
\end{figure}

There is tremendous value in humanoid robots taking on the dull, dirty, and dangerous work in spaces built for humans.
However, an autonomy system must coordinate locomotion, whole-body motion, perception, contact, and operator supervision.
It must also support adaptation to new tasks.
The central claim of this work is that our behavior autonomy architecture enables fast, resilient, and adaptive humanoid robot behaviors.
Different autonomy architectural choices affect how quickly a behavior executes, how robustly it tolerates task variation and disturbance, and how much effort is required to adapt an existing behavior to a new variant.
This work focuses on the runtime structure that makes them faster, more resilient, and easier to modify than previously published systems~\cite{Johnson_2017,calvert2024behavior}.
This work demonstrates its relevance through performant real-robot demos on a variety of humanoid robot platforms, including IHMC's recently developed Alex, a custom, fully electric humanoid robot with 29 degrees of freedom, shown in \autoref{fig:20260309_AlexRightPullDoorProfessionalStill}.
This robot uses PSYONIC Ability Hands~\cite{psyonic_ability_hand}, which are anthropomorphic 5-finger hands with 6 degrees of freedom each.
It also perceives the world onboard, using just two stereo color cameras in the head with a human-like interpupillary distance.

We have run versions of the presented system with several humanoid robots, including the Boston Dynamics DARPA Robotics Challenge Finals-Era Atlas, Unitree's H1-2, and IHMC's Alex, reflecting the generality of our approach.

\begin{figure*}[t]
    \centering
    \includegraphics[width=\textwidth]{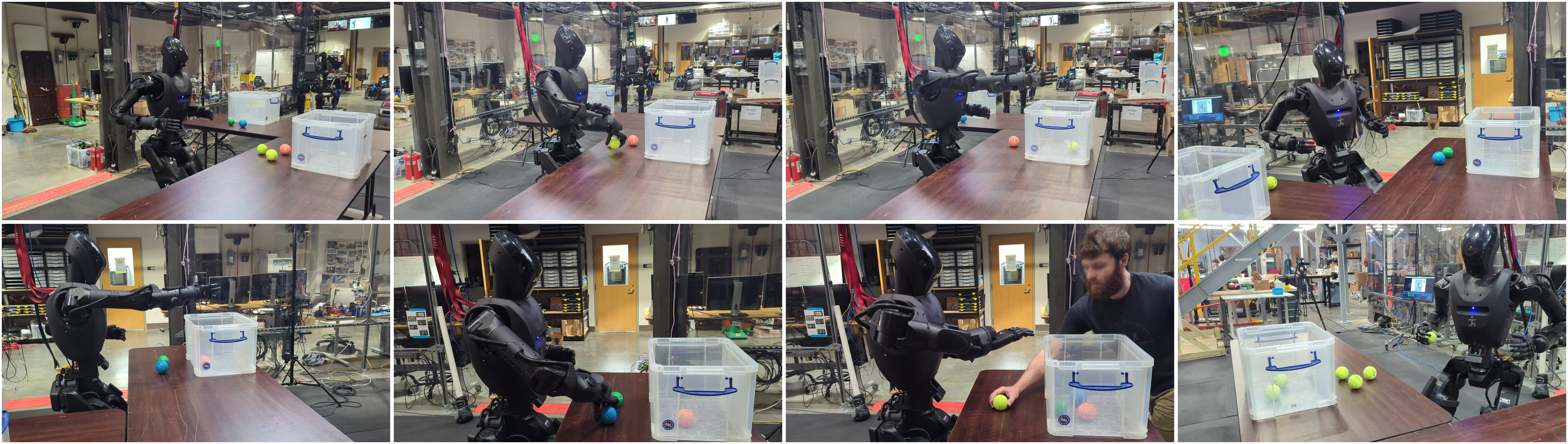}
    \caption[Alex two-table ball sorting.]{Alex performing a two-table loco-manipulation reactive ball-sorting task.
    The robot sorted nine balls across two tables in 2 minutes and 8 seconds.
    Supplementary video material accompanies this submission.}
    \label{fig:AlexLocoBallSortingTwoTables}
\end{figure*}

\subsection{Problem Statement and Scope}

This paper addresses robot-local behavior authoring and execution for humanoid loco-manipulation in human-scale environments, where the robot must walk, reach, manipulate objects, perceive doors and tables, and respond to operator input without external tracking infrastructure.

The work presented here focuses on a behavior architecture that unifies task execution and perception.
In this architecture, behaviors are authored as reusable structures that can be executed on the robot and inspected and modified at runtime.
This design can improve task speed, robustness under variation, and the time required to adapt a behavior to a new task variant such as a door or station.

We compare the proposed system to prior systems reported in the literature and to published comparative results.
However, we do not experimentally reproduce results from the literature.
We also do not experimentally evaluate off-the-shelf alternatives such as MoveIt~\cite{MoveIt_2019}, MoveIt Pro~\cite{moveit_pro}, BehaviorTree.CPP~\cite{behaviortree_cpp}, and Groot~\cite{groot1, groot2}.

This work aims to address three primary questions:
\begin{enumerate}
    \item What concrete behavior architecture is sufficient to enable fast and robust performance across a range of humanoid loco-manipulation behaviors?
    \item How does this architecture compare with prior IHMC baselines and reported reinforcement learning door systems on overlapping metrics such as traversal time, reliability, and task variation coverage?
    \item How does runtime-editable behavior structure change the time and sequence of steps required for an expert operator to create new behaviors and adapt, extend, and compose existing ones?
\end{enumerate}
To answer these questions, we establish the following hypotheses, which are essentially the characteristics of the architecture we committed to building.
\begin{enumerate}
    \item Robot-local execution with synchronized UI state, concurrent action layering, reactive tree logic, and behavior-time semantic perception yield door behaviors that are faster and more reliable than prior IHMC baselines and competitive with reported reinforcement learning systems on overlapping door tasks.
    \item Runtime-editable behaviors and perception modules reduce the iteration loop required to diagnose failures, modify logic, and re-test on the robot, relative to redeploy, restart, or retrain workflows.
    \item Decomposing behaviors into reusable primitives, subtrees, and scene actions allows new door and loco-manipulation variants to be brought up by editing a small part of a working behavior rather than rebuilding it from scratch.
\end{enumerate}

We organize the evaluation around three lenses---Speed, Resilience, and Adaptability---corresponding to execution time, robustness under disturbance and task variation, and runtime editability for task retargeting.

The main contributions of this paper are:
\begin{itemize}
    \item \textbf{Architecture:} A robot-local behavior architecture that unifies runtime-editable task logic, synchronized operator UI state, and behavior-time perception for humanoid loco-manipulation, evaluated through repeated real-robot demonstrations on multiple humanoid platforms, with Alex as the primary evaluation robot.
    \item \textbf{Speed:} Humanoid door traversals among the fastest timed results in the published literature, with performance competitive with recent learned humanoid door policies on overlapping tasks.
    \item \textbf{Combined evaluation:} To our knowledge, the first door-traversal study to report competitive speed and repeated-trial reliability on a humanoid while also measuring behavior authoring and adaptation time on the same runtime-editable stack.
    \item \textbf{Adaptability:} Evidence that an expert operator can bring a novel humanoid door behavior from an empty sequence to first fully autonomous success in under two hours of measured active authoring time, with comparable adaptation durations for retargeting existing behaviors to new doors and tasks.
    \item \textbf{Perception:} Behavior-time perception modules for generalized door traversal and table approach, including visual door-state estimation across diverse door configurations and a depth-based table-edge detection method that yields a reusable approach frame for precise humanoid positioning.
\end{itemize}

The rest of this paper is organized as follows.
\autoref{ch:prior_work} reviews related work on reusable action structure, reactive task coordination, human-machine coordination, perception, and real robot door task demonstrations.
\autoref{ch:architecture2} describes the behavior architecture, covering the runtime-editable tree, the synchronized operator interface, and the behavior-time perception scene.
\autoref{ch:defense} presents the real robot evaluation organized around speed, resilience, and adaptability, and compares the results to prior systems from the literature.
\autoref{ch:pontification} discusses the strong and weak points of the system and outlines next steps.

\subsection{Door Traversals as a Benchmark Task}

We use door traversal as a benchmark task because it exposes the full coordination problem in a compact and repeatable setting.
There are three main phases of door traversal: the approach, the opening, and the traversal walk.
For the approach, footsteps must be precise to provide arm reachability while avoiding collisions between the door and the robot.
Since the doors often have spring closers, a complex manipulation sequence is required to open the door all the way without failure.
Finally, the traversal walk is difficult because the robot must maintain balance while avoiding collisions with the door frame and resisting lateral impacts from the spring-loaded door panel.
\autoref{fig:20260309_AlexRightPullDoorProfessionalStill} shows Alex executing a right pull lever-handle door traversal with PSYONIC Ability Hands.

Although doors are the primary benchmark, the architecture is intended for loco-manipulation more broadly.
In \autoref{ch:defense}, we present six real robot task variants developed during this work.
In one demonstration, the robot walks between two tables and sorts colored balls into the correct containers, as shown in \autoref{fig:AlexLocoBallSortingTwoTables}.

\section{Related Work}
\label{ch:prior_work}
In this section we will cover literature that relates to this work on reusable action structure, reactive task coordination, human-machine coordination, perception, and real robot door task demonstrations.
Architectural references include CLARAty~\cite{Volpe_2001_CLARAty}, Affordance Templates~\cite{Hart_2014, Hart_2015}, Affordance Primitives~\cite{Pettinger_2020, Pettinger_2022}, Coactive Design~\cite{Johnson_2014}, Director~\cite{Marion_2017}, FlexBE~\cite{Schillinger_2016}, RAFCON~\cite{Brunner_2016}, Drawing Board~\cite{Senft_2021}, and the CENTAURO behavior tree systems~\cite{2023_centaur_bt, 2024_wang_grounded_lm}.
Door task references span from classical and model-based mobile-manipulator and humanoid systems to recent learned legged and humanoid policies.
The systems with metrics that overlap our benchmark are compared with each other and this work in \autoref{ch:defense}.

\subsection{Architectural Foundations}

Affordance Templates, Behavior Trees, and Coactive Design form the strongest theoretical foundations of this work.
Affordance Templates define an architecture for reusable behavior with respect to recurring tasks in the environment.
Behavior Trees provide a data structure for organizing and orchestrating behavior.
Coactive Design poses three questions that drive the design of a system to leverage the synergistic opportunities of human-robot teams.

\subsubsection{Affordance Templates}
\label{sec:ats}

The Affordance Template Framework~\cite{Hart_2014, Hart_2015} provides a way to parameterize and reuse robot loco-manipulation behaviors with respect to environmental affordances.
The word ``affordance'' comes from James Gibson's 1979 book, \emph{The Ecological Approach to Visual Perception}~\cite{Gibson_1979}.
It is used in the sense that an environmental feature ``affords'' an action.
For example, the ground is an affordance for walking and a handle is an affordance for turning.
An affordance template is a robotics construct for defining and parameterizing robot behavior with respect to environmental affordances.
This makes it one of the most consequential design inspirations for our work.

The framework was developed and used for the DARPA Robotics Challenge Trials in 2013, while affordance templates were first presented in the literature in 2014.
Early work on NASA's Valkyrie humanoid robot demonstrated a valve turning task and included templates for door opening, walking, hose mating, valve turning, and ladder and stair climbing~\cite{Hart_2014}.
In 2022, Hart et al.~\cite{Hart_2022} generalized and further demonstrated the power of this approach.
Tools for footstep planning, motion planning, stance generation, and grasping were incorporated to enable increased autonomous functionality.
This was demonstrated on NASA Valkyrie for integrated task execution, including car-door interaction and explosive-device handling~\cite{2019_Jorgensen_valEOD}.
In that work, the operator could load, edit, and execute templates at runtime.

Related works on affordance primitives~\cite{Pettinger_2020, Pettinger_2022} emphasize constrained interaction motions such as turning a valve or closing a drawer.
The screw primitive in particular shows how a reusable low-level action can capture a class of constrained manipulation motions without hard-coding them.

\subsubsection{Behavior Trees}

Behavior Trees~\cite{2018_colledanchise, 2022_iovino_behavior_trees} are a data structure that coordinates low-level actions through structured and reactive task logic.
They define logical operator nodes such as sequence, fallback, and parallel which work to control the execution flow of a behavior.
Actions are performed by the leaf nodes which command the robot and gather environmental data.
They provide reactivity through a ``ticking'' system in which each tick starts at the root node.
This is in contrast to state machines in which each tick starts from the previous state.
By continuously re-evaluating from the top down, from tick to tick there are pathways to ending up in very different parts of the tree without explicit connections between those parts.

A Behavior Tree is ticked from the root node at a constant frequency.
The tick is a signal which propagates through the tree, causing nodes to evaluate or execute.
Control nodes redirect the tick and condition and action nodes return results: success, failure, or running.
A node is executed if and only if it receives ticks.
If an action is underway, it returns running to the parent.
When it is done it returns success or failure.
A sequence node is the most fundamental control node.
It runs each node in order while they are successful.
A fallback node runs each child in order while earlier children fail.
The remaining node types include parallel composition, conditions, actions, and decorators.
\subsubsection{Coactive Design}

The Coactive Design method~\cite{Johnson_2014} is an iterative process comprised of three main subprocesses: identification, selection and implementation, and evaluation of change.
The most complex is the identification process, in which requirements, alternatives, and interdependence relationships are explored.
A set of desired interdependence relationships are determined and selected for implementation.
The result is then evaluated using human feedback and performance analysis.
This methodology was used for the design and development of Team IHMC's operator interface in the 2015 DRC.
A later analysis details how that methodology led to success in the competition~\cite{Johnson_2017}.

Coactive Design treats human-machine systems as interdependent and emphasizes the value in making the system observable, predictable, and directable, with interdependence analysis charts used to document those relationships.
The architecture in this work is meant to be developed, debugged, and adapted by expert operators on real hardware.
Coactive Design directly motivates the authoring and supervision interfaces in \autoref{ch:architecture2}.
Our prior work~\cite{calvert2024behavior} documents an interdependence analysis for a related runtime-editable behavior architecture.

\subsubsection{Door Traversal Systems}

Prior door-traversal systems span behavior decompositions, coupled base-arm planning, and learned whole-body policies.
Classical mobile-manipulator work includes Jain and Kemp~\cite{Jain2008dooropening,Jain2010EPC}, who decompose push- and pull-side opening on statically stable platforms with force-based contact and externally cued handles; Chitta et al.~\cite{2010ChittaDoorOpening}, who plan coupled PR2 base--arm motion after grasp acquisition; Axelrod and Huang~\cite{Axelrod_2015}, who execute semi-autonomous PackBot push/pull knob and lever traversal under operator door classification; and Banerjee et al.~\cite{Banerjee_2015}, who demonstrate supervised Atlas door traversal for the DRC.

Recent planner- and behavior-centric systems extend to full sequences and legged bases: Jang et al.~\cite{2023JangDoorTraversal} use graph search to plan a Husky--Franka through approach, opening, traversal, and closing; Sleiman et al.~\cite{Sleiman_2023} solve ANYmal spring-loaded door traversal with bilevel task and motion planning; Thamrongaphichartkul and Vongbunyong~\cite{Thamrongaphichartkul_2024} author ROS~2 behavior-tree traversal on a differential-drive manipulator in simulation; Kang et al.~\cite{kang2024door} integrate segmentation, force-based direction identification, and adaptive or RL opening on a wheeled manipulator; and Schulze et al.~\cite{Schulze_2025} deploy a Kinova-equipped SCITOS for building navigation including doors.

Learned systems target faster open-and-traverse execution: Zhang et al.~\cite{zhang2024learningopentraversedoors} train a single ANYmal teacher--student policy for push and pull doors with external doorway measurements; Xue et al.~\cite{xue2025opening} present DoorMan, a humanoid RGB sim-to-real policy; and Zhang et al.~\cite{Zhang_2026_Sumo} steer a pretrained RL whole-body controller with sample-based MPC, with humanoid door opening demonstrated in simulation.

Our work is unique among these references in combining robot-local execution, runtime-editable structure, behavior-time perception, and real-world demonstration on a humanoid robot.
DoorMan is competitive but runs vision inference off-board, and the learned door systems concentrate on policy retraining rather than runtime structural edits.
Relative to wheeled manipulators and quadrupeds, humanoid door traversal additionally requires concurrent management of foot placement, bipedal balance, reachability, collision avoidance, and bimanual hand contacts~\cite{Johnson_2017}.

\section{System Architecture}
\label{ch:architecture2}
In this section, we will cover the architecture of our behavior system.
Our design choices center on enabling the utilization of expert human knowledge in creating autonomous behaviors on humanoid robots.
We do this by including a human operator in the loop through tight synchronization between a user interface and the robot.
Once behaviors are set up sufficiently, our architecture allows the robot to function autonomously by keeping all perception and control on board the robot.

\subsection{Domain}
\begin{figure*}[t]
    \noindent\hspace*{-0.13\textwidth}\resizebox{\textwidth}{!}{\input{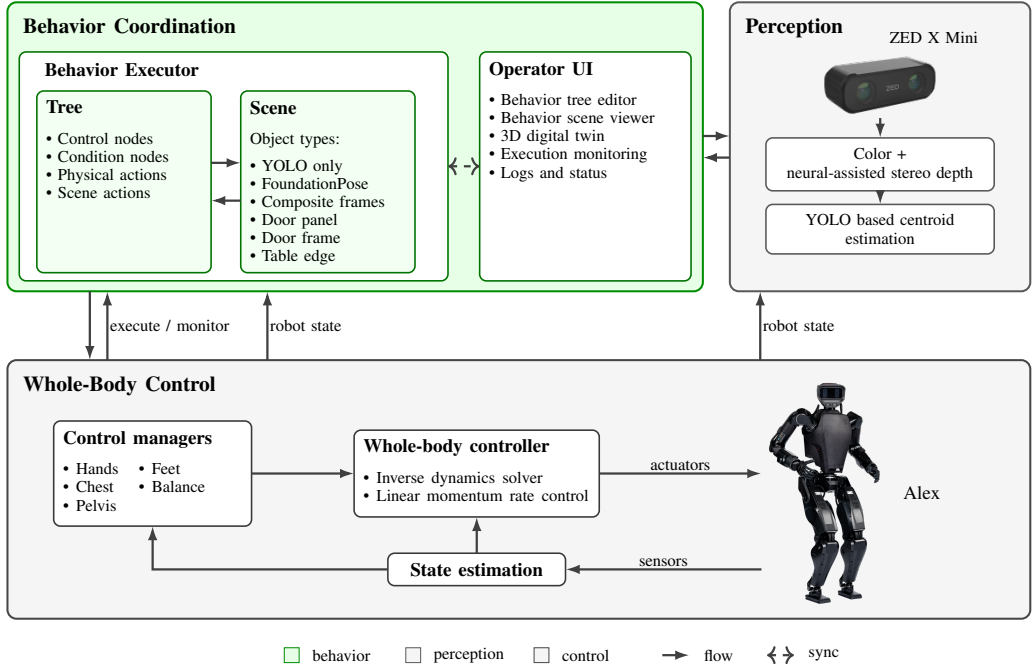}}
    \caption[Domain of the behavior system.]{%
        This figure illustrates the domain of our behavior system.
        Our work centers on behavior coordination and how it integrates with perception and whole-body control.
        Our behavior executor owns and manages a tree and scene in coordination with a human operator.
        Perception is provided via a ZED X Mini~\cite{zed_x_mini} and semantic object detections from YOLO~\cite{redmon2016yolo}.
        Physical behavior actions are executed via a whole-body controller which can achieve objectives asynchronously and simultaneously.
        Control objectives include moving or locking parts of the body and walking.}
    \label{fig:Behavior_Overview}
\end{figure*}

There are many perspectives from which to view our system.
\autoref{fig:Behavior_Overview} summarizes the primary components by domain: Behavior Coordination, Perception, and Whole-Body Control.
This highlights that our system is built separately from depth and semantic perception and from whole-body control.
Our work sits firmly in the top-left area and focuses on how it interfaces with the others.
The behavior executor owns the behavior tree and a behavior-specific scene.
The operator UI owns the behavior editor, 3D digital robot twin, and system monitoring.
We use lowercase ``behavior tree'' for our implementation and capitalized ``Behavior Tree'' when referring to the literature formalism~\cite{2018_colledanchise}.

We use a ZED X Mini, a stereo color camera pair with human-like interpupillary distance that computes high-fidelity, neural-assisted depth from the stereo images.
A key dependency of the behavior system is YOLO~\cite{redmon2016yolo}, an algorithm that provides high-frequency semantic object detection and instance segmentation.
It is crucial to our door behaviors and loco-manipulation behaviors, which perceive the world onboard with no external sensing or fiducial markers.

The other key dependency is a whole-body controller that can reliably walk, pose, and exert forces on the world.
Ours is set up uniquely to accept asynchronous commands for footsteps and the different body parts, combining those concurrent requests into whole-body motions while balancing~\cite{Koolen_2016}

\subsection{Process Structure}

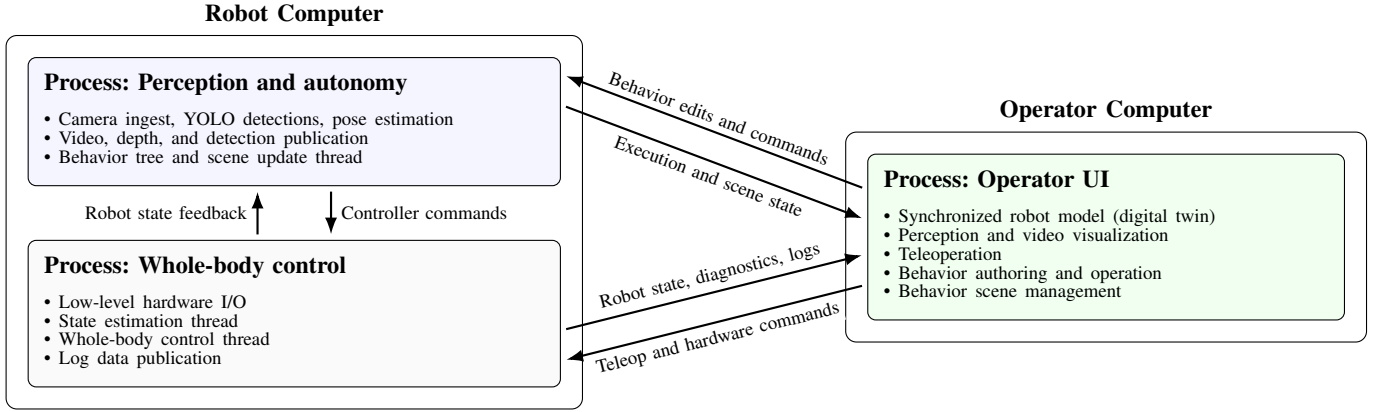
\begin{figure*}[t]
\centering
    \resizebox{\textwidth}{!}{    \begin{tikzpicture}[
        font=\small,
    proc/.style={draw, rounded corners=3pt, align=left, inner sep=6pt, text width=0.36\textwidth, fill=blue!4},
    procgray/.style={draw, rounded corners=3pt, align=left, inner sep=6pt, text width=0.36\textwidth, fill=gray!4},
    op/.style={draw, rounded corners=3pt, align=left, inner sep=6pt, text width=0.32\textwidth, fill=green!6},
    flow/.style={
        -{Latex[length=2.2mm,width=1.6mm]}, thick,
        shorten >=2pt, shorten <=2pt
    },
    flowlabel/.style={
        sloped, midway, fill=white, inner sep=1.5pt,
        font=\scriptsize
    },
    vlabel/.style={
        midway, fill=white, inner sep=1.5pt, font=\scriptsize
    },
    group/.style={draw, rounded corners=4pt, inner sep=8pt}
        ]
        \node[proc] (autonomy) at (0, 0) {
            \textbf{Process: Perception and autonomy}\\[3pt]
            {\scriptsize
            \begin{tabular}[t]{@{}l@{}}
                \textbullet\ Camera ingest, YOLO detections, pose estimation\\[-1pt]
                \textbullet\ Video, depth, and detection publication\\[-1pt]
                \textbullet\ Behavior tree and scene update thread
            \end{tabular}}
        };

        \node[procgray, below=0.7cm of autonomy] (control) {
            \textbf{Process: Whole-body control}\\[3pt]
            {\scriptsize
            \begin{tabular}[t]{@{}l@{}}
                \textbullet\ Low-level hardware I/O\\[-1pt]
                \textbullet\ State estimation thread\\[-1pt]
                \textbullet\ Whole-body control thread\\[-1pt]
                \textbullet\ Log data publication
            \end{tabular}}
        };

        \node[group, fit=(control) (autonomy), label={[font=\small]above:\textbf{Robot Computer}}] (robotgroup) {};

        \node[op, right=4.0cm of control, yshift=1.0cm] (operator) {
            \textbf{Process: Operator UI}\\[3pt]
            {\scriptsize
            \begin{tabular}[t]{@{}>{\raggedright\arraybackslash}p{\linewidth}@{}}
                \textbullet\ Synchronized robot model (digital twin)\\[-1pt]
                \textbullet\ Perception and video visualization\\[-1pt]
                \textbullet\ Teleoperation\\[-1pt]
                \textbullet\ Behavior authoring and operation\\[-1pt]
                \textbullet\ Behavior scene management
            \end{tabular}}
        };

        \node[group, fit=(operator), label={[font=\small]above:\textbf{Operator Computer}}] (operatorgroup) {};

        \def\robotLinkX{0.48cm}
        \def\linkSep{0.20cm}
        \def\autonomyLinkY{0.42cm}
        \def\controlLinkY{-0.42cm}

        \draw[flow] ([xshift=\robotLinkX]autonomy.south) --
            node[vlabel, right=2pt]{Controller commands}
            ([xshift=\robotLinkX]control.north);
        \draw[flow] ([xshift=-\robotLinkX]control.north) --
            node[vlabel, left=2pt]{Robot state feedback}
            ([xshift=-\robotLinkX]autonomy.south);

        \draw[flow] ([yshift={\autonomyLinkY+\linkSep}]operator.west) --
            node[flowlabel, above=1pt]{Behavior edits and commands}
            ([yshift={\autonomyLinkY+\linkSep}]autonomy.east);
        \draw[flow] ([yshift={\autonomyLinkY-\linkSep}]autonomy.east) --
            node[flowlabel, below=1pt]{Execution and scene state}
            ([yshift={\autonomyLinkY-\linkSep}]operator.west);

        \draw[flow] ([yshift={\controlLinkY+\linkSep}]control.east) --
            node[flowlabel, above=1pt]{Robot state, diagnostics, logs}
            ([yshift={\controlLinkY+\linkSep}]operator.west);
        \draw[flow] ([yshift={\controlLinkY-\linkSep}]operator.west) --
            node[flowlabel, below=1pt]{Teleop and hardware commands}
            ([yshift={\controlLinkY-\linkSep}]control.east);
    \end{tikzpicture}}
    \caption[Runtime structure.]{%
        The runtime structure used in this work.
        Two processes run on the robot, and an operator process supports authoring, supervision, and visualization.}
    \label{fig:RuntimeStructureAlex}
\end{figure*}

In \autoref{fig:RuntimeStructureAlex} we show another view of the system, where the parts are grouped by process and location.
This is meant to illustrate that the behavior system and perception run on board the robot in a process alongside the control process, whereas the operator UI is dislocated from the robot.
This structure enables tight integration between the perception, behavior, and control stack on the robot for both performance and autonomy.
This allows high-rate color, depth, and scene data to stay local with low latency between processes.
The operator UI inspects, edits, pauses, single-steps, and triggers autonomous execution but does not own task progression.

The decoupling of the user interface and the on-robot stack matters operationally.
If the UI crashes or communications degrade, the runtime keeps the state needed to continue the current task or stop in a controlled way according to its authored logic.
It also matters during development.
The operator can reconnect, inspect the same execution, and resume working without reconstructing behavior state from an external script or a lost UI session.
The split also isolates responsibilities cleanly: autonomy and perception focus on decision-making and scene state, control focuses on real-time balance, walking, and hardware I/O, and the operator process can be rich and inspectable without sitting in the critical execution path.

\subsection{Node Structure}

\begin{figure*}[t]
\centering
\newsavebox{\behaviorNodeDecompBox}
\sbox{\behaviorNodeDecompBox}{%
    \resizebox{0.52\textwidth}{!}{
\ifdefined\TIKZWEBINCLUDE
    \setlength{\textwidth}{469.75502pt}
    \setlength{\linewidth}{469.75502pt}
\fi
    \newcommand{\nodetitle}[1]{{\small\bfseries #1}}
    \newcommand{\nodecode}[1]{{\ttfamily\footnotesize #1}}
    \begin{tikzpicture}[
        font=\small,
        wrapperbox/.style={
            font={\small\setstretch{1.0}},
            draw, rounded corners=3pt, align=left, inner sep=5pt,
            text width=0.34\textwidth, minimum height=2.6cm
        },
        sharedbox/.style={
            font={\small\setstretch{1.0}},
            draw, rounded corners=3pt, align=left, inner sep=5pt,
            text width=0.34\textwidth, minimum height=1.9cm
        },
        ui/.style={wrapperbox, fill=green!8},
        exec/.style={wrapperbox, fill=blue!6},
        state/.style={sharedbox, fill=orange!8},
        definition/.style={sharedbox, fill=gray!12},
        owns/.style={-{Diamond[open=false, length=2.6mm, width=2.0mm]}, thick,
        shorten >=2pt, shorten <=2pt},
        colheader/.style={font=\fontsize{11.5}{13.8}\selectfont, fill=white, inner sep=1pt}
    ]
    \def\colx{3.3}
    \def\statey{-3.6}
    \def\defy{-6.5}
    \def\headery{1.62}

    \node[colheader] at ( \colx,  \headery) {\textbf{Operator Process}};
    \node[colheader] at (-\colx,  \headery) {\textbf{Robot Process}};

    \node[ui] (rdx) at (\colx, 0) {%
        \nodetitle{User Interface}\\
        \nodecode{RDXBehaviorTreeNode}\\[2pt]
        \begin{itemize}
        \item Only in the authoring UI process
        \item Renders editing widgets and 3D visualization
        \end{itemize}};
    \node[exec] (execn) at (-\colx, 0) {%
        \nodetitle{Executor}\\
        \nodecode{BehaviorTreeNodeExecutor}\\[2pt]
        \begin{itemize}
        \item Only in the on-robot autonomy process
        \item Sole caller into the whole-body controller
        \end{itemize}};

    \node[state] (state) at (0, \statey) {%
        \nodetitle{Runtime State \normalfont\textit{(synchronized across processes)}}\\
        \nodecode{BehaviorTreeNodeState}\\[2pt]
        \begin{itemize}
        \item Node ID, active status, recent log messages
        \end{itemize}};

    \node[definition] (def) at (0, \defy) {%
        \nodetitle{Persistent Definition \normalfont\textit{(JSON on disk)}}\\
        \nodecode{BehaviorTreeNodeDefinition}\\[2pt]
        \begin{itemize}
        \item Name, notes, parameters, children
        \end{itemize}};

    \draw[owns] (rdx.south)   -- ([xshift= 1.5cm]state.north);
    \draw[owns] (execn.south) -- ([xshift=-1.5cm]state.north);
    \draw[owns] (state.south) -- (def.north);

    \node[font=\scriptsize, anchor=north] at (0, \defy - 1.45)
        {\tikz[baseline=-0.6ex]\draw[owns] (0,0) -- (2mm,0);~owns};
    \end{tikzpicture}}%
}
\begin{minipage}[c]{\wd\behaviorNodeDecompBox}
    \usebox{\behaviorNodeDecompBox}
\end{minipage}%
\hspace{0.02\textwidth}%
\begin{minipage}[c][\dimexpr\ht\behaviorNodeDecompBox+\dp\behaviorNodeDecompBox\relax][c]{0.43\textwidth}
    \input{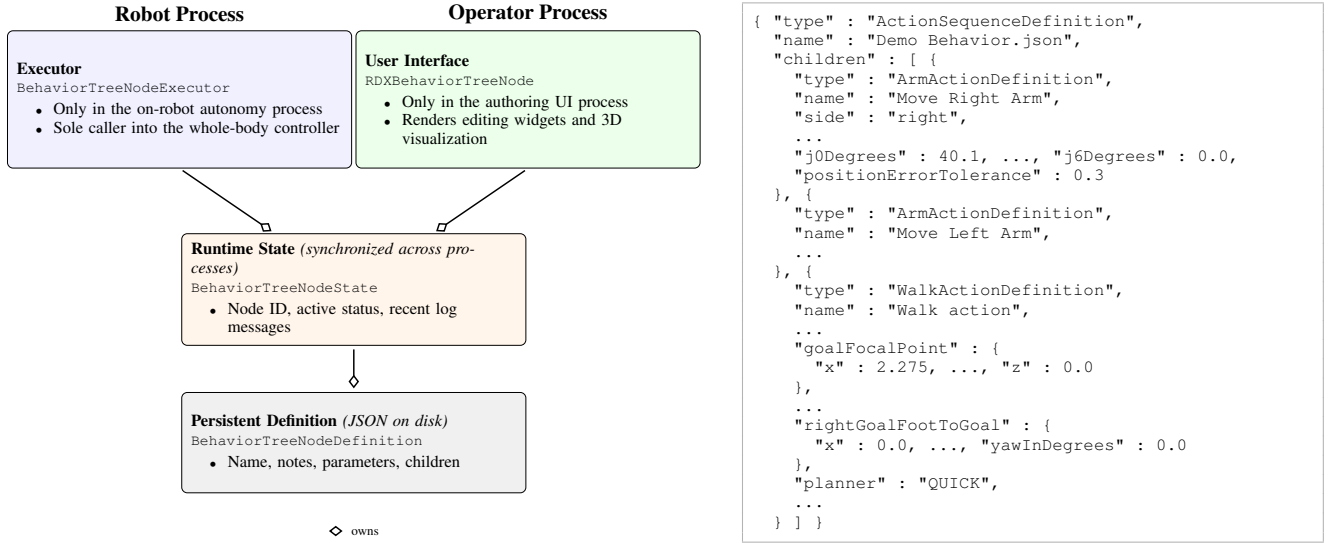}%
\end{minipage}
\caption[Four-part decomposition used for each behavior-tree node.]{%
    Four-part decomposition used for each behavior-tree node (left).
    Two process-specific layers, UI and executor, each own an instance of a runtime state, which in turn owns a persistent definition.
    The synchronization layer keeps the runtime state and the live-edited portion of the definition consistent across processes, while the full definition is persisted as JSON.
    The condensed JSON excerpt (right) illustrates the \texttt{type}, \texttt{children}, and parameter fields saved for each node.}
\label{fig:BehaviorNodeDecomposition}
\end{figure*}

Each behavior tree node is decomposed into four parts as shown in \autoref{fig:BehaviorNodeDecomposition}: a robot-side executor, an operator-side UI, a synchronizing runtime state, and a persistent definition.
The persistent definition stores authored content (name, notes, parameters, children).
The runtime state instantiates the definition and adds runtime information (node identity, active status, recent log messages).
The UI and execution implementations wrap the state and definition layers, allowing the separation by process while sharing synchronized runtime state and authored structure.
Operator interaction, on-robot execution, synchronized state, and definition are each isolated in separate code files.

\subsection{Robot-Operator Data Synchronization}

\begin{figure*}[t]
\centering
    \resizebox{\textwidth}{!}{\providecommand{\syncbullet}{\textbullet\ }
\newcommand{\synchead}[1]{{\footnotesize\bfseries #1}}
\newcommand{\synccode}[1]{{\ttfamily\scriptsize #1}}
\begin{tikzpicture}[
    font=\small,
    panel/.style={
        draw, rounded corners=4pt, align=left, inner sep=6pt, line width=0.8pt
    },
    sidepanel/.style={
        panel, text width=4.45cm
    },
    op/.style={sidepanel, draw=green!45!black, fill=green!8},
    robot/.style={sidepanel, draw=blue!45!black, fill=blue!8},
    mid/.style={panel, text width=5.20cm, draw=black!45, fill=gray!10},
    aux/.style={panel, text width=7.35cm, draw=orange!70!black, fill=orange!10},
    groupline/.style={draw=black!45, rounded corners=5pt, line width=0.9pt},
    localflow/.style={<->, line width=0.9pt, draw=black!70},
    cmdflow/.style={-{Latex[length=2.2mm,width=1.5mm]}, line width=1.0pt, draw=green!50!black},
    statusflow/.style={-{Latex[length=2.2mm,width=1.5mm]}, line width=1.0pt, draw=blue!55!black},
    clockflow/.style={<->, dashed, line width=0.95pt, draw=orange!70!black},
    grouphead/.style={font=\small\bfseries},
    note/.style={font=\scriptsize, align=center, fill=white, inner sep=1pt}
]

\node[op] (opTree) at (-6.1, 1.55) {%
{\synchead{Local Tree Instance}}\\[-1pt]
{\synccode{RDXBehaviorTreeNode + BehaviorTree}}\\[4pt]
{\scriptsize
\begin{tabular}[t]{@{}l@{}}
\syncbullet authored edits\\[-1pt]
\syncbullet mirrored robot and task info
\end{tabular}}};

\node[robot] (robotTree) at (6.1, 1.55) {%
{\synchead{Local Tree Instance}}\\[-1pt]
{\synccode{BehaviorTreeNodeExecutor + BehaviorTree}}\\[4pt]
{\scriptsize
\begin{tabular}[t]{@{}l@{}}
\syncbullet executor and scene logic\\[-1pt]
\syncbullet task state remains robot-local
\end{tabular}}};

\node[op] (opSync) at (-6.1, -1.55) {%
{\synchead{Per-Tick Wrapper}}\\[-1pt]
{\synccode{ROS2BehaviorTree}}\\[4pt]
{\scriptsize
\begin{tabular}[t]{@{}l@{}}
\syncbullet \texttt{updateSubscription()}\\[-1pt]
\syncbullet local UI tick\\[-1pt]
\syncbullet \texttt{updatePublication()}
\end{tabular}}};

\node[robot] (robotSync) at (6.1, -1.55) {%
{\synchead{Per-Tick Wrapper}}\\[-1pt]
{\synccode{ROS2BehaviorTree}}\\[4pt]
{\scriptsize
\begin{tabular}[t]{@{}l@{}}
\syncbullet \texttt{updateSubscription()}\\[-1pt]
\syncbullet local executor tick\\[-1pt]
\syncbullet \texttt{updatePublication()}
\end{tabular}}};

\node[mid] (message) at (0, -1.55) {%
{\synchead{30 Hz Tree Snapshot}}\\[-1pt]
{\synccode{BehaviorTreeStateMessage}}\\[4pt]
{\scriptsize
\begin{tabular}[t]{@{}l@{}}
\syncbullet sequence ID and next node ID\\[-1pt]
\syncbullet root and topology modification records\\[-1pt]
\syncbullet depth-first typed snapshot\\[-1pt]
\syncbullet full or partial node payloads
\end{tabular}}};

\node[aux] (clock) at (0, -4.75) {%
{\synchead{Clock-Offset Support for CRDT Merge}}\\[-1pt]
{\synccode{ROS2PeerClockOffsetEstimator}}\\[-1pt]
{\synccode{ROS2PeerClockOffsetEstimatorPeer}}\\[4pt]
{\scriptsize
\begin{tabular}[t]{@{}l@{}}
\syncbullet 5 Hz ping/reply between publisher GUIDs\\[-1pt]
\syncbullet half round-trip time estimates peer clock offset
\end{tabular}}};

\node[groupline, draw=green!45!black, fit=(opTree) (opSync), inner sep=9pt] (opGroup) {};
\node[groupline, draw=blue!45!black, fit=(robotTree) (robotSync), inner sep=9pt] (robotGroup) {};
\node[grouphead, anchor=south] at (opGroup.north) {Operator UI Process};
\node[grouphead, anchor=south] at (robotGroup.north) {Robot Autonomy Process};

\draw[localflow] (opTree.south) -- (opSync.north);
\draw[localflow] (robotTree.south) -- (robotSync.north);

\draw[cmdflow] ([yshift=0.52cm]opSync.east) -- ([yshift=0.52cm]message.west);
\draw[cmdflow] ([yshift=0.52cm]message.east) -- ([yshift=0.52cm]robotSync.west);
\draw[statusflow] ([yshift=-0.52cm]robotSync.west) -- ([yshift=-0.52cm]message.east);
\draw[statusflow] ([yshift=-0.52cm]message.west) -- ([yshift=-0.52cm]opSync.east);

\draw[clockflow] ([xshift=1.10cm]opSync.south) -- ([xshift=-2.30cm]clock.north);
\draw[clockflow] ([xshift=-1.10cm]robotSync.south) -- ([xshift=2.30cm]clock.north);

\end{tikzpicture}}
    \caption[Behavior-tree synchronization between the operator UI and robot executor.]{%
        Behavior-tree synchronization between the operator UI and robot executor.
        Each side maintains a local tree instance and a per-tick synchronization wrapper.
        A 30~Hz depth-first tree snapshot carries either full or partial per-node payloads.
        Bidirectional CRDT fields use clock-offset-corrected last-writer arbitration; robot-owned status fields are mirrored one way to the UI.}
    \label{fig:BehaviorTreeSynchronization}
\end{figure*}
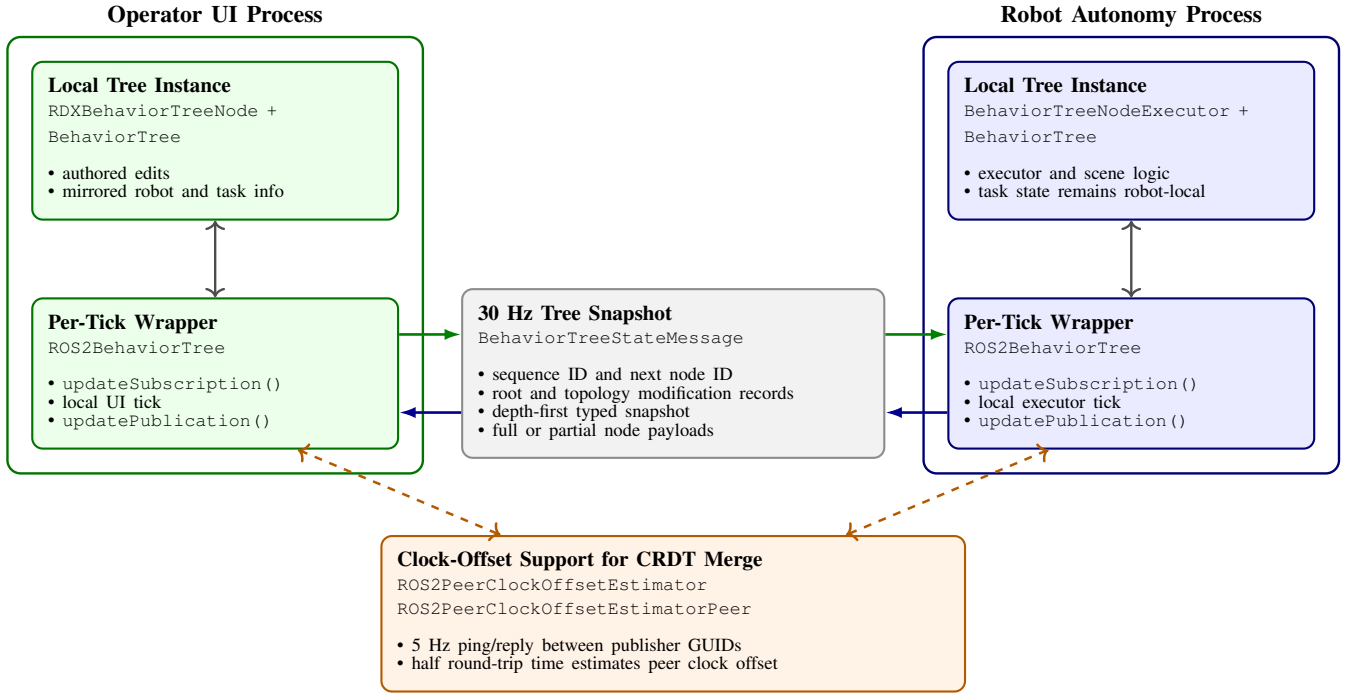

The cost of our process structure is that it necessitates a complex synchronization mechanism between the robot and the operator to facilitate rich runtime editability.
We formulate our behavior tree and scene state as Conflict-Free Replicated Data Types (CRDTs)~\cite{shapiro2011crdt} and synchronize them at 30~Hz, as shown in \autoref{fig:BehaviorTreeSynchronization}.

We use the ROS 2~\cite{ros2} DDS middleware to transmit the synchronization messages. Its autodiscovery feature streamlines the connections between operator UIs and robots.
The robot and the operator are allowed to concurrently edit data types such as footstep goal poses, which may be moved by the robot to update their pose with respect to the parent frame while the operator may wish to modify the definition of those goal poses.
We use a latest-timestamp algorithm to resolve potential conflicts.
The latest modification to a data field persists.
A clock-offset estimator, shown at the bottom of \autoref{fig:BehaviorTreeSynchronization}, supports this design by providing a means for comparing data modification times.

The operator interface and the robot-side behavior system symmetrically synchronize on each tick.
At the start of each tick, the wrappers apply any received data updates.
After the UI or executor performs its local work, it republishes the current tree and increments an update number that is used to maintain order.

The serializer packs the tree in depth-first order using a node type table and per-type message arrays.
To save bandwidth, each node is sent as a compact partial-data payload unless a full type-specific payload is needed.
Full data are sent when the node definition has changed, when a peer has requested retransmission, or when the node reports fresh status.
The compact version carries only the generic state and CRDT metadata.
This mechanism is summarized in \autoref{fig:BehaviorTreeSynchronization}.

Topology is synchronized separately from per-node content.
The tree root reference, whole-tree metadata including the next available node ID, and each node's child list each carry their own modification metadata.
On receipt, the subscriber reconstructs an intermediate message tree, matches nodes by ID, and applies topology operations only where the incoming root or child-list modification is newer.
A topology change queue is used to apply all topology changes at once, later, to ensure tree consistency.
New nodes are replicated locally through the node builder, moved or reordered nodes are reattached through the topology queue, and nodes that disappear from the incoming tree are destroyed.
If a process comes online late or a dropped message leaves only partial data available, the receiver flags the node as needing full data and the peer resends the full payload on a later publication.

Some node fields, such as execution status and visualization data, are unidirectional and only modifiable by one actor.
Examples include leaf execution information such as ``is next for execution'', ``can execute'', ``is executing'', and ``has failed'', which are owned by the robot side.
These are mirrored to the other side without arbitration.
Other fields are concurrently modifiable by all actors, which we refer to as bidirectional fields.
Examples include the root node's ``automatic execution'' gate, ``next execution index'' selection, manual-step requests, ``enable concurrency'', preview mode, and ``reset failures''.

The bidirectional CRDT fields use a last-modified record that stores author identity, a monotonic modification number, and a timestamp.
A local write records the writer GUID, modification number, and timestamp.
On receipt, a newer modification number wins immediately.
Equal-number races are resolved by comparing timestamps after transforming the peer timestamp into the local clock frame using a peer clock-offset estimator.
This estimator derives the time offset from ROS 2 ping-reply messages under a symmetric-delay assumption.

\subsection{Persistence Storage of Behaviors}

To enable authored structure to be saved, loaded, copied, and versioned, we support saving to and loading from JSON files.
The saved JSON represents the definition layer from \autoref{fig:BehaviorNodeDecomposition}.
It contains names, notes, parameter values, child hierarchy, and node type specifics.
Each node is serialized with a \texttt{type} field and a \texttt{children} array that mirrors the behavior tree.
A condensed excerpt appears at the right of \autoref{fig:BehaviorNodeDecomposition}, illustrating nested arm and walk action definitions.

We have tried to make these files readable by humans, but they are not intended to be edited.
For example, instead of saving a number with unnecessarily high precision (i.e.\ ``0.1349159123''), we round it (i.e.\ ``0.135'').
We also use degrees instead of radians because it is easier to reason about.
Still, we recommend viewing behaviors in the operator interface for the best understanding.

\subsection{Whole-Body Controller}

The behavior coordination layer interfaces with a whole-body walking controller that executes footsteps and trajectory requests for body parts.
The behavior runtime sends requests to the controller and monitors execution through status messages.
Our implementation is based on IHMC's linear momentum rate control module and whole-body inverse dynamics solver.
The control objective manager accepts asynchronous commands for footsteps and upper-body tasks, combining concurrent requests into balanced whole-body motions including walking.
A more complete description of the controller we base our system on, including its architectural layout, QP formulation, and walking modules, can be found in~\cite{calvert2024behavior}.

\subsection{Operator Interface}

Our operator interface is based on a robotics software graphics framework called Robot Data eXplorer (RDX).
It provides a collection of tools needed to interact with robot data, featuring dockable panels of widgets, a 3D scene with interactable elements, and VR support.
It makes it easy to build domain-specific applications for not only the operator interface, but development UIs for perception and planning.
It is built over Dear ImGui, libGDX, and OpenVR~\cite{dearimgui,libgdx,openvr}, together with the IHMC libraries Euclid~\cite{ihmc_euclid} and Mecano~\cite{ihmc_mecano}.
RDX is used for behavior authoring, digital twin visualization, and direct teleoperation modes such as whole-body streaming, footstep placement, and joystick walking.

\subsection{Object-Centric Action Definition}

In our system, physical actions can be defined in the coordinate frame of a scene object or a part of the robot.
This includes hand poses, footsteps, spine action, the screw primitive axes, and the neck action.
When approaching objects like tables and doors, footsteps are defined with respect to the perceived object or door frame.
When grasping an object, the pre-grasp hand poses are defined with respect to the perceived object.
This important architectural design element is inspired by both the IHMC DARPA Robotics Challenge user interface~\cite{Johnson_2017} and the Affordance Template Framework~\cite{Hart_2014, Hart_2015} discussed in Section~\ref{sec:ats}.

This is also referred to as ``object-centric'' action, which is acting with respect to the object, versus ``ego-centric'' action, which is acting with respect to a robot-local frame such as the pelvis.
Object-centric action definition is essential and core to the reusability of behaviors.
It serves two main purposes.
The first is to enable varied starting conditions of the robot.
For example, task approach should be specified with respect to the task.
The other is to reduce compounding errors.
That is because long sequences of actions defined in a fixed world frame are subject to state estimator drift and control inaccuracy, whereas actions defined relative to a currently perceived object can re-anchor each step.

If a door traversal behavior was executed fully ego-centrically, assuming the robot started at the same initial pose as during authoring, it would likely miss the handle grasp or run into the door panel or frame.
As an example of why, the door traversal approach stance may not achieve the desired footsteps accurately due to bad balance and recovery during the final steps.
This would result in a stance that is slightly different than planned.
If it were to then try to grasp the handle with respect to that stance (self) instead of the handle, even though it may have worked before, it will likely not work this time.
The pre-grasp hand pose would be misaligned.
Instead, by using the actively perceived or re-perceived door handle pose to define the pre-grasp action, the grasp should succeed, overcoming the prior error using the inverse kinematics solver to achieve the correct pose.

Object-centric action definition can be viewed as a practical form of sequential composition in the sense proposed by~\cite{1999_burridge}.
One interpretation is that successive object-anchored actions act like Lyapunov funnels, where task approach is funnel A\@ and the handle grasp is funnel B\@.
The opening of funnel A is the space of possible starting locations of the robot.
The size and shape of the opening of funnel B would represent the tolerance to different robot stance locations, mainly based on arm reachability.
Under this reading, object-centric actions help maintain task stability despite accruing state and control errors.

\begin{figure*}[t]
    \centering
    \includegraphics[width=1.6\columnwidth]{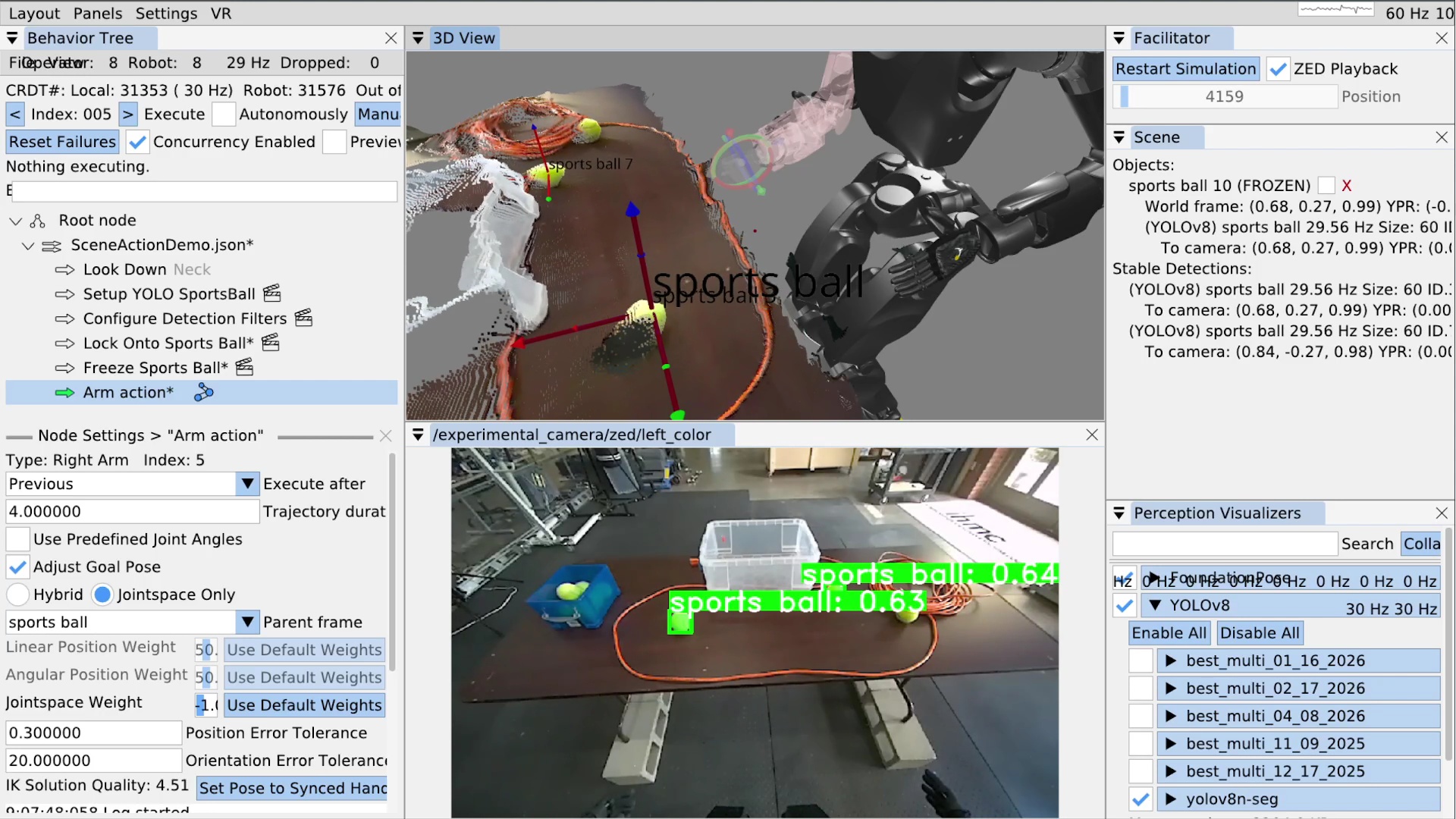}
    \caption{A screenshot of the operator interface.
    On the mid-left, the behavior tree editor is shown with neck and scene actions that acquire a lock on a sports ball.
    In the bottom left, parameters for authoring an arm action are shown, with ``sports ball'' as the selected frame.
    In the center, a red, green, and blue 3D pose gizmo is used to author the hand pose relative to the ball.
    In the top right, the current behavior scene items are shown.}
    \label{fig:guide_sim_scene_action_arm}
\end{figure*}

The frame-based action math is implemented using the Euclid library~\cite{ihmc_euclid}.
We instantiate a reference frame tree with world frame as the root and the robot and all scene objects as subtrees and leaves.
Euclid makes it easy to create new reference frames and calculate transformations between any two frames in the tree.
\autoref{fig:guide_sim_scene_action_arm} shows the operator interface when authoring an object-centric action.

\subsection{Behavior Tree Structure}

Our behaviors are structured as a tree of nodes.
The tree organization serves a dual-purpose.
It allows for organization of behavior and also a way to author logic in a simple way.

We reuse the concept of a filesystem in computing for behaviors.
Robot behaviors can be abstracted across many layers.
Useful tasks for robot work are defined as completable goals which usually consist of higher-level abstractions like ``sort the objects into containers'' or ``deliver this package to room 40''.
These high level task specifications ultimately need to be composed as a sequence of low-level primitive actions.
High level task specifications live towards the root of the tree and low level primitive actions live as the leaves.

\begin{figure}[!b]
\centering
    \small
    \resizebox{\columnwidth}{!}{\begin{tikzpicture}[
    node distance=1.4cm and 1.6cm,
    every node/.style={font=\sffamily},
    btnode/.style={
        draw,
        minimum width=1.5cm,
        minimum height=1.0cm,
        align=center,
        fill=white,
        thick
    },
    actionnode/.style={
        btnode,
        minimum width=2.2cm
    },
    condition/.style={
        ellipse,
        draw,
        thick,
        fill=white,
        align=center,
        inner xsep=6pt,
        inner ysep=3pt,
        minimum width=2.4cm
    },
    flow/.style={
        thick
    },
    nextexec/.style={
        ->,
        very thick,
        green!60!black
    },
    gotoflow/.style={
        ->,
        thick,
        dashed,
        draw=black!65
    },
    successflow/.style={
        ->,
        thick,
        dotted,
        draw=black!65
    },
    leafindex/.style={
        draw,
        line width=0.9pt,
        fill=white,
        font=\scriptsize\bfseries,
        inner sep=1.8pt,
        rounded corners=1.5pt
    },
    fbseqbox/.style={
        draw,
        dashed,
        thick,
        rounded corners=4pt,
        inner sep=5pt
    },
    flowlabel/.style={
        font=\small,
        fill=white,
        inner sep=1.5pt
    },
    captionlabel/.style={
        font=\small
    },
    btdecornode/.style={
        draw,
        thick,
        fill=white,
        align=center,
        font=\sffamily,
        text width=1.45cm,
        minimum width=1.55cm,
        minimum height=1.28cm,
        inner xsep=5pt,
        inner ysep=3pt
    }
]

\newcommand{\btrdecorsym}[1]{%
    \begingroup
    \setbox0=\hbox{?}%
    \raisebox{-\dp0}{\hbox to \wd0{\hfil\raisebox{0pt}[\ht0][0pt]{#1}\hfil}}%
    \endgroup
}

\newcommand{\btrdecor}[2]{%
    \btrdecorsym{#1}\\[2pt]%
    {\small #2}%
}

\newcommand{\btrrootsym}{%
    \btrdecorsym{\scalebox{1.25}[1]{$\emptyset$}}%
}

\newcommand{\btrseqsym}{%
    \begingroup
    \setbox0=\hbox{\raisebox{-0.02ex}{\scalebox{1.15}{$\rightarrow$}}}%
    \raisebox{-0.02ex}{\hbox to 0pt{\hss\kern-0.15\wd0\box0\hss}}%
    \endgroup
}

\newcommand{\btrleafindexrect}[2]{%
    \node[leafindex, anchor=north west]
        at ([xshift=0.8pt,yshift=-0.8pt]#1.north west) {#2};%
}

\newcommand{\btrleafindexoval}[2]{%
    \node[leafindex, anchor=north west, opacity=0, name=-btridxmeas] at ([xshift=0.8pt,yshift=-0.8pt]#1.155) {#2};%
    \path let \p1=(-btridxmeas.south west), \p2=(-btridxmeas.south east),
        \n1={\x2-\x1} in
        node[leafindex, anchor=north west]
            at ([xshift={-3*\n1}] -btridxmeas.north west) {#2};%
}

\node[btdecornode] (root) {\btrdecor{\btrrootsym}{Root}};

\node[btdecornode, below=1.2cm of root] (seq) {\btrdecor{\btrseqsym}{Sequence}};

\node[actionnode, below left=1.5cm and 3.4cm of seq] (actionA) {Action A};
\node[btdecornode, below=1.5cm of seq] (fb) {\btrdecor{?}{Fallback}};
\node[actionnode, below right=1.5cm and 3.4cm of seq] (actionB) {Action B};

\node[actionnode, below=2.0cm of fb] (catchAct) {Catch action};
\node[condition, left=0.8cm of catchAct] (tryCond) {Try condition};
\node[actionnode, right=0.85cm of catchAct] (gotoAct) {Goto action};

\draw[flow] (root) -- (seq);
\draw[flow] (seq) -- (actionA);
\draw[flow] (seq) -- (fb);
\draw[flow] (seq) -- (actionB);
\draw[flow] (fb) -- (tryCond);
\draw[flow] (fb) -- (catchAct);
\draw[flow] (fb) -- (gotoAct);

\coordinate (execarrowstart) at ([xshift=-1.1cm]actionA.west);

\draw[nextexec] (execarrowstart) -- (actionA.west);

\draw[gotoflow]
    (gotoAct.north west)
    .. controls +(-2.0cm, 1.8cm) and +(3.2cm, -1.5cm) ..
    (actionA.south east)
    node[pos=0.82, flowlabel, sloped, above]
        {Retry};

\draw[successflow]
    (tryCond.north east)
    .. controls +(2.2cm, 1.6cm) and +(-1.4cm, -1.0cm) ..
    (actionB.south west)
    node[pos=0.75, flowlabel, sloped, above]
        {Success};

\node[fbseqbox, fit=(catchAct)(gotoAct)] (fbseq) {};
\node[captionlabel, below=3pt of fbseq.south] {Fallback sequence};

\btrleafindexrect{actionA}{0}
\btrleafindexoval{tryCond}{1}
\btrleafindexrect{catchAct}{2}
\btrleafindexrect{gotoAct}{3}
\btrleafindexrect{actionB}{4}

\end{tikzpicture}}
    \caption[Representative sample behavior tree.]{%
        A sample tree illustrating our sequence and fallback functionality.
        The next action to execute is held as state between updates, as indicated by the green arrow pointing to Action A.
        In our tree, ``everything is a sequence'' in the depth-first ordering of the leaves, which are numbered in the figure.
        When the fallback try fails, the rest of the children are executed as a sequence.
        When it succeeds, the catch sequence is skipped.
        Goto nodes are used to re-route execution manually.}
    \label{fig:BehaviorTreeStructureSample}
\end{figure}
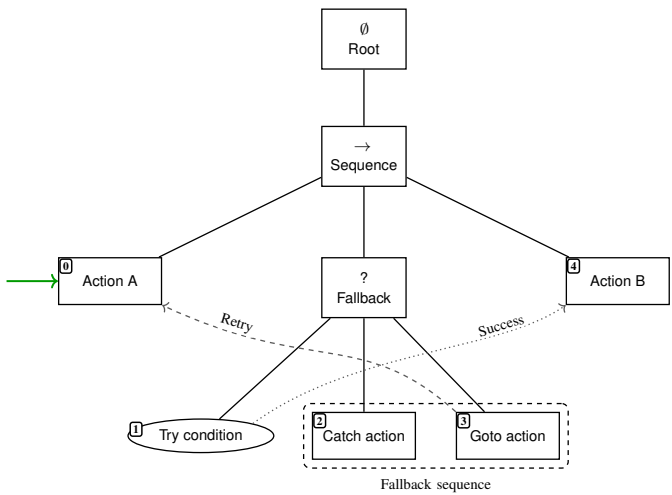

Our system is not a Behavior Tree as in the literature~\cite{2018_colledanchise} (we will use capitalized Behavior Tree to reference the literature version).
It is rather a tree of behavior implemented in a way we found most useful.
However, we do adopt Behavior Trees' concepts of sequence, fallback, condition, and action nodes.
\autoref{fig:BehaviorTreeStructureSample} summarizes the main structural elements.

Our runtime updates all nodes on each tick and does not reduce control flow to a fixed set of return types such as ``success,'' ``failure,'' and ``running.''
The node library is freer in how it represents and exposes execution state, and the model stays closer to how we actually authored and debugged humanoid door traversals, where action overlap, retries, manual stepping, and scene-state inspection mattered more than adherence to a standard BT interface.

In our implementation, the whole tree is one big sequence.
The order of the sequence is the depth-first ordering of the leaves.
The basic sequence algorithm runs the depth-first leaf order on each tick.

A unique part of our architecture, which diverges even further from Behavior Trees, is that we keep a next execution index.
This is in contrast to restarting at the root node on each tick.
In this way, we are more like a state machine.
The next execution index is a pointer to the next node to execute, which can be changed by the operator, the sequence executor, a goto node, or a fallback node.

The flow of execution in our system treats the whole tree as a big sequence with the leaves being the executable elements in depth-first ordering.
Each leaf node can be triggered for execution and holds ``is executing'' and ``has failed'' states.
Concurrency is supported throughout the tree via ``execute after'' dependencies, as described in~\cite{calvert2024behavior}.
When the behavior is executing in ``autonomous mode'', actions are not triggered if the node pointed to via the ``execute after'' field is executing.
This way of defining concurrency provides flexibility to the behavior author without being overly constraining.
Arrows are rendered in the UI to help the operator identify the defined concurrency at a glance.

We also have a unique implementation of a fallback node, which is not a list of steps to try, but rather one thing to try and a sequence as a catch.
A fallback node is shown in \autoref{fig:BehaviorTreeStructureSample}.
The fallback node also supports concurrency.
The first ``concurrent group'' is treated as the ``try'' and the ``catch'' will not execute until all try nodes have completed.
This is necessary so the try can be evaluated and the future execution path can be determined.

The tree view in the operator interface is closer to a hierarchical file browser than to a dense 2D graph editor.
That choice matched the rest of the operator workflow because the behavior view had to coexist with a 3D scene, first-person sensor imagery, scene-object panels, robot-status panels, and teleoperation widgets.
As an example, the behavior tree editor abstracts the authored structure of our generalized door traversal behavior as an expandable entry called \texttt{door/Door\-Traversal.json}.
When expanded, its children are shown, expandable themselves, including door behavior subtypes such as \texttt{door/Right\-Pull\-Door.json}.

In summary, our implementation is neither a Behavior Tree, a state machine, nor a hierarchical state machine.
Instead, it more closely resembles a computer program, combining elements from each to suit the needs of a runtime-editable behavior authoring system.

\subsection{Node Library}

Our node library includes the following node types:

\begin{itemize}
    \item \textbf{Sequence.} Organizational node, since the entire tree is treated as a sequence. Can be used to define subsequences.
    \item \textbf{Fallback.} A two-part subtree with a ``try'' and a ``catch''. If the try succeeds, the catch is skipped.
    \item \textbf{Condition.} A leaf node that computes success or failure based on some criteria. An example is checking for spatial occupancy in a point cloud.
    \item \textbf{Goto.} Holds an operator-defined or dynamically computed reference to a node to goto next. When executed, the node to goto is selected as next for execution.
    \item \textbf{Checkpoint.} A no-op action with a name that holds semantic meaning to the operator. Can be used to aid in concurrent action scheduling.
    \item \textbf{Scene Action.} Performs an action on the behavior scene. This includes setting up, freezing, or deleting scene objects and configuring the perception models such as YOLO and FoundationPose~\cite{wen2024foundationpose}.
    \item \textbf{Door Traversal.} A sequence node with built-in heuristics for generalized door traversal. It routes the flow of execution to the appropriate door behavior type based on detected door parameters by the door frame scene object.
    \item \textbf{Neck Action.} Commands a neck roll-pitch-yaw trajectory.
    \item \textbf{Spine Action.} Commands a jointspace or taskspace spine trajectory.
    \item \textbf{Walk Action.} Holds a set of pre-authored footsteps or goal steps for active footstep planning. Commands the footsteps to the robot when executed.
    \item \textbf{Arm Action.} Constructs and commands the left or right arm joints via jointspace or taskspace goals. Fails when the achieved hand pose is not within a configurable error tolerance.
    \item \textbf{Screw Primitive Action.} An arm action that is parameterized by a helix. The helix is generated from an object-centric axis and the current hand pose. This motion is useful for turning handles and opening doors.
    \item \textbf{Pelvis Action.} Commands a pelvis pose trajectory in taskspace.
    \item \textbf{Ability Hand Action.} Commands the six degree-of-freedom PSYONIC Ability Hand using preset grips or operator defined joint angles.
    \item \textbf{Wait Action.} Waits for the defined duration and then completes.
    \item \textbf{Leg Action.} Commands a single leg, held in the air, in taskspace or jointspace while balancing on the other leg.
\end{itemize}

\subsection{Shape-Contains Condition Node}

A key reactive element in our system is our condition node that checks point and color containment.
It provides a semantic-free perceptual capability that operates only from the depth point cloud, making it robust.
It uses a CUDA kernel to count the number of points and average color within a behavior-frame-relative virtual shape, such as a sphere or a capsule.
Per-node authorable thresholds specify a minimum and maximum number of points and optionally minimum and maximum Hue-Saturation-Value that define a successful result.
The result can be used in a sequence to halt the behavior on failure or in a fallback ``try'' to branch the flow of execution based on the result.

\begin{figure}[t]
    \centering
    \includegraphics[width=0.9\columnwidth]{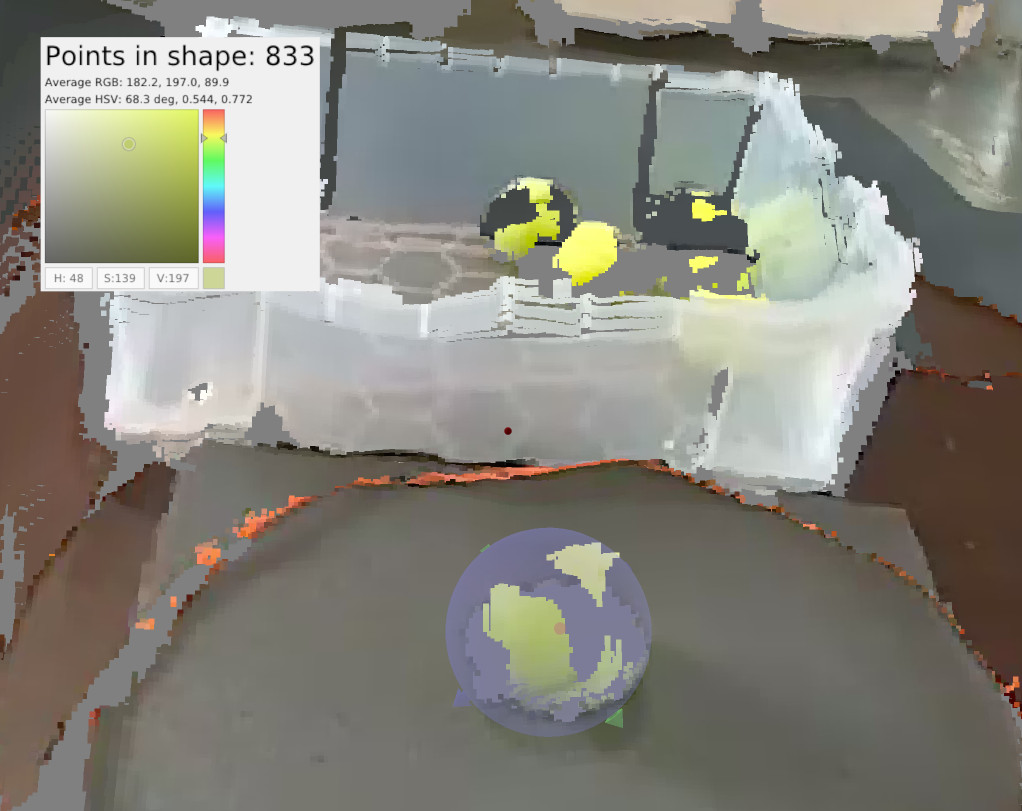}
    \caption{A GPU-accelerated kernel counts points and average color within a virtual sphere.
    Here, it is detecting 833 yellow points for a tennis ball.
    This mechanism is used by our condition node to provide authorable perceptual reactivity.}
    \label{fig:shape_contains_condition}
\end{figure}

The core detection mechanism is demonstrated in \autoref{fig:shape_contains_condition}.
The check typically runs in under 10~ms, which makes it fast enough to evaluate inside the behavior execution loop without incurring visible pauses.
Since this mechanism depends only on the depth and color images, it is highly robust in contrast to semantic detectors such as YOLO.
In our demonstrations, we used this mechanism to detect door panels when opened, obstacles such as humans in the doorway, and the presence and color of tennis balls when held in the robot's hand.

\subsection{Behavior Scene}

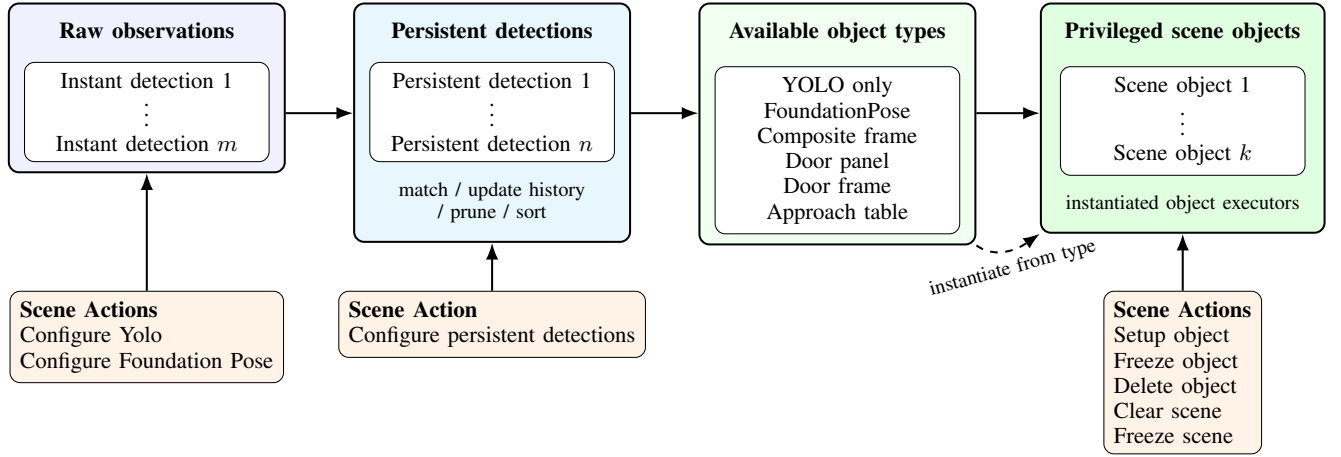
\begin{figure*}[!t]
\centering
    \resizebox{\textwidth}{!}{\usetikzlibrary{backgrounds,positioning,calc,fit}
\begin{tikzpicture}[
    font=\small,
    >=Latex,
    node distance=6mm and 8mm,
    box/.style={draw, rounded corners, align=center, inner sep=4pt},
    area/.style={draw, rounded corners, thick, inner sep=4pt},
    flow/.style={->, thick},
    action/.style={draw, rounded corners, fill=black!5, align=left, inner sep=4pt},
    listbox/.style={draw, rounded corners, align=center, inner sep=4pt, fill=white, minimum width=34mm},
    title/.style={align=center, font=\small\bfseries},
    area pad/.style={inner xsep=6pt, inner ysep=4pt}
]

\coordinate (col0) at (0, 0);
\coordinate (col1) at (48mm, 0);
\coordinate (col2) at (96mm, 0);
\coordinate (col3) at (144mm, 0);

    \node[title, anchor=north] (raw-title) at (col0) {Raw observations};

    \node[listbox, anchor=north] (rawlist) at ([yshift=-2mm]raw-title.south)
        {Instant detection 1\\
    $\vdots$\\
    Instant detection $m$};

    \node[title, anchor=north] (persist-title) at (col1) {Persistent detections};

    \node[listbox, anchor=north] (plist) at ([yshift=-2mm]persist-title.south)
        {Persistent detection 1\\
    $\vdots$\\
    Persistent detection $n$};

    \node[align=center, font=\footnotesize, anchor=north] (pmatch) at ([yshift=-1.5mm]plist.south)
        {match / update history\\ / prune / sort};

    \node[title, anchor=north] (types-title) at (col2) {Available object types};

    \node[listbox, anchor=north] (typelist) at ([yshift=-2mm]types-title.south)
        {YOLO only\\
    FoundationPose\\
    Composite frame\\
    Door panel\\
    Door frame\\
    Approach table};

    \node[title, anchor=north] (objs-title) at (col3) {Privileged scene objects};

    \node[listbox, anchor=north] (objlist) at ([yshift=-2mm]objs-title.south)
        {Scene object 1\\
    $\vdots$\\
    Scene object $k$};

    \node[align=center, font=\footnotesize, anchor=north] (oexec) at ([yshift=-1.5mm]objlist.south)
        {instantiated object executors};

    \begin{scope}[on background layer]
        \node[area, fill=blue!6, area pad, fit=(raw-title)(rawlist)] (raw) {};
        \node[area, fill=cyan!8, area pad, fit=(persist-title)(plist)(pmatch)] (persist) {};
        \node[area, fill=green!6, area pad, fit=(types-title)(typelist)] (types) {};
        \node[area, fill=green!12, area pad, fit=(objs-title)(objlist)(oexec)] (objs) {};
    \end{scope}

    \path let \p1=(raw.south), \p2=(persist.south), \p3=(types.south), \p4=(objs.south),
        \n1={min(\y1,\y2,\y3,\y4)} in coordinate (pipeline-bottom) at (0, \n1);

    \node[action, fill=orange!10, anchor=north, minimum width=34mm] (rawact)
        at ([yshift=-6.3mm]pipeline-bottom -| raw.south)
    {
        \textbf{Scene Actions}\\
        Configure Yolo\\
        Configure Foundation Pose
    };

    \node[action, fill=orange!10, anchor=north, minimum width=34mm] (pconfig)
        at ([yshift=-6.3mm]pipeline-bottom -| persist.south)
    {
        \textbf{Scene Action}\\
        Configure persistent detections
    };

    \node[action, fill=orange!10, anchor=north] (objsceneact)
        at ([yshift=-6.3mm]pipeline-bottom -| objs.south)
    {
        \textbf{Scene Actions}\\
        Setup object\\
        Freeze object\\
        Delete object\\
        Clear scene\\
        Freeze scene
    };

    \path let \p1=(raw.center), \p2=(persist.center), \p3=(types.center), \p4=(objs.center),
        \n1={(\y1+\y2+\y3+\y4)/4} in coordinate (flow-y) at (0, \n1);

    \draw[flow] (raw.east |- flow-y) -- (persist.west |- flow-y);
    \draw[flow] (persist.east |- flow-y) -- (types.west |- flow-y);
    \draw[flow] (types.east |- flow-y) -- (objs.west |- flow-y);

    \draw[flow] (rawact.north) -- (raw.south);
    \draw[flow] (pconfig.north) -- (persist.south);
    \draw[flow] (objsceneact.north) -- (objs.south);

    \draw[flow, dashed] (types.south east) to[out=-35,in=-145]
        node[below, sloped, font=\footnotesize] {instantiate from type} (objs.south west);

\end{tikzpicture}}
    \caption[Behavior scene overview.]{%
        The behavior scene consists of a list of active persistent detections and a privileged list of objects.
        Object types can be implemented to derive structured information about the scene.
        Scene actions are used to manage the privileged objects and the underlying perception models.}
    \label{fig:BehaviorScene}
\end{figure*}

The last major architectural component of our system is the behavior scene.
It consists of two areas: an actively updated list of tracked detections and a behavior-maintained, privileged list that actions can reference.
These lists are part of the behavior system and for use only by the behaviors.
This exclusivity is a core design choice: behaviors retain full control over the perception they depend on, rather than sharing a global world model with unrelated modules.
That separation is especially important given the technical limitations of robot perception systems, where detection quality, latency, and occlusion vary by task and viewpoint.

Since leveraging perception to accomplish tasks is difficult, we leave the creativity up to the human operator.
To do this, we have a library of scene actions which may be placed in the behavior tree and executed in the same way as the physical actions like moving a hand.
\autoref{fig:BehaviorScene} shows where the scene actions affect the behavior scene perception pipeline.
It also lists the current set of object and scene action types.

Our scene object types include directly detected object types, such as YOLO-only and FoundationPose objects, but also derived types that compute new frames from persistent detections, existing scene objects, or depth data.
These specialized scene object types are a demonstration of how advanced perception and world modeling can be incorporated into the system to extend capability to novel tasks.
In the next several sections, we will describe these derived types.

\subsection{Door Panel Object}

Our door panel object is a derived scene object built from two stable YOLO persistent detections: one for the opening mechanism and one for the door panel.
A setup object type scene action can be used to create a door panel object after those detections have passed the usual stability and proximity checks.

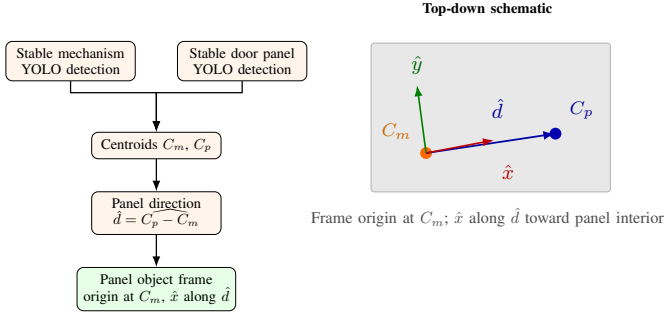
\begin{figure}[h]
\centering
    \resizebox{\columnwidth}{!}{\begin{tikzpicture}[
    font=\small,
    >=Latex,
    node distance=5mm and 7mm,
    flow/.style={->, thick},
    stepbox/.style={draw, rounded corners, align=center, inner sep=4pt, minimum width=28mm, fill=orange!8},
    outbox/.style={draw, rounded corners, align=center, inner sep=4pt, minimum width=30mm, fill=green!10},
    title/.style={font=\small\bfseries, align=center},
    note/.style={font=\footnotesize, align=center, text=black!70}
]

\node[stepbox] (mech) at (-2.4, 3.85) {Stable mechanism\\YOLO detection};
\node[stepbox, right=9mm of mech] (panel) {Stable door panel\\YOLO detection};
\node[stepbox, below=11mm of $(mech.south)!0.5!(panel.south)$] (centroids)
    {Centroids $C_m$, $C_p$};
\node[stepbox, below=7mm of centroids] (line)
    {Panel direction\\$\hat{d} = \widehat{C_p - C_m}$};
\node[outbox, below=7mm of line] (frame)
    {Panel object frame\\origin at $C_m$, $\hat{x}$ along $\hat{d}$};

\draw[flow] (mech.south) -- ++(0,-2.5mm) -| (centroids.north);
\draw[flow] (panel.south) -- ++(0,-2.5mm) -| (centroids.north);
\draw[flow] (centroids) -- (line);
\draw[flow] (line) -- (frame);

\begin{scope}[shift={(4.325, 1.15)}]
    \node[title, anchor=south] at (2.30, 3.55) {Top-down schematic};

    \begin{scope}[shift={(2.30,1.55)}, scale=1.4, every node/.append style={scale=1.4}, shift={(-2.30,-1.55)}]
        \fill[gray!18, draw=gray!55, line width=0.7pt, rounded corners=2pt]
            (0.50, 0.35) rectangle (4.20, 2.65);
        \coordinate (Cm) at (1.35, 0.95);
        \fill[orange!80!red] (Cm) circle (2.6pt);
        \node[font=\footnotesize, anchor=south east, text=orange!80!black] at ($(Cm)+(-1mm,1mm)$) {$C_m$};

        \coordinate (Cp) at (3.35, 1.25);
        \fill[blue!70!black] (Cp) circle (2.6pt);
        \node[font=\footnotesize, anchor=south west, text=blue!60!black] at ($(Cp)+(1mm,1mm)$) {$C_p$};

        \draw[blue!65!black, line width=1.1pt, -{Latex[length=2.6mm]}] (Cm) -- (Cp);
        \node[font=\small, anchor=south, text=blue!60!black] at ($(Cm)!0.55!(Cp)+(0,2.5mm)$)
            {$\hat{d}$};

        \draw[red!75!black, line width=1.0pt, -{Latex[length=2.4mm]}] (Cm) -- ++(1.05, 0.20);
        \draw[green!50!black, line width=1.0pt, -{Latex[length=2.4mm]}] (Cm) -- ++(-0.14, 1.05);
        \node[font=\small, text=red!70!black, anchor=north west] at ($(Cm)+(1.05,-0.05)$) {$\hat{x}$};
        \node[font=\small, text=green!40!black, anchor=south] at ($(Cm)+(-0.14,1.10)$) {$\hat{y}$};
    \end{scope}

    \node[font=\normalsize, align=center, text=black!70, anchor=north] at (2.30, -0.35)
        {Frame origin at $C_m$; $\hat{x}$ along $\hat{d}$ toward panel interior};
\end{scope}

\end{tikzpicture}}
    \caption[Door panel object algorithm.]{%
        The door panel object algorithm.
        Stable YOLO persistent detections for the opening mechanism and door panel yield centroids $C_m$ and $C_p$.
        The vector from $C_m$ to $C_p$ defines panel orientation under the vertical hinge assumption.
        The privileged panel object frame is placed at $C_m$ with $\hat{x}$ pointing toward the panel interior.}
    \label{fig:DoorPanelObjectAlgorithm}
\end{figure}

The object executor draws a line between the mechanism centroid and the panel centroid and uses that direction to define panel orientation.
This exploits the assumption that door panels swing on a vertical hinge.
The resulting frame is centered on the mechanism and establishes the mechanism side and the panel interior direction.
\autoref{fig:DoorPanelObjectAlgorithm} summarizes the computation and the resulting top-down frame convention.
Subsequent actions can then use this frame for grasp, unlatch, and opening motions.

\subsection{Door Frame Object}

The door frame object extends the door panel object with semantic door geometry estimated from the live depth image.
A door frame setup object type scene action requires an existing door panel object to be established in the scene.
It then derives the door frame pose, push or pull type, and door open angle from that panel state and the current point cloud.

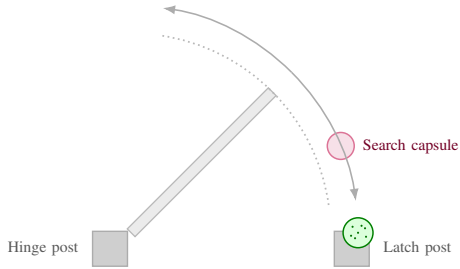
\begin{figure}[h]
\centering
    \resizebox{0.7\columnwidth}{!}{\begin{tikzpicture}[
    font=\small,
    >=Latex,
    post/.style={fill=gray!38, draw=gray!65, line width=0.8pt},
    panel/.style={fill=gray!16, draw=gray!55, line width=0.75pt},
    sweep/.style={dotted, line width=1pt, gray!55},
    searchcap/.style={draw=purple!55, fill=purple!12, line width=0.85pt},
    detectcap/.style={draw=green!50!black, fill=green!14, line width=0.85pt},
    sweeparrow/.style={->, line width=0.9pt, gray!65}
]

\def\panelW{4.0}
\def\panelT{0.22}
\def\openAngle{45}
\def\outerR{4.55}
\def\postSize{7mm}
\def\postGap{0.06}

\coordinate (H) at (0, 0);

\node[post, minimum size=\postSize, inner sep=0pt, anchor=south east] (hingepost)
    at (-\postGap, 0.42-7mm) {};
\node[post, minimum size=\postSize, inner sep=0pt, anchor=south west] (latchpost)
    at (\panelW+\postGap, 0.42-7mm) {};
\node[font=\small, anchor=east, text=black!65] at ($(hingepost.west)+(-1.5mm,0)$)
    {Hinge post};
\node[font=\small, anchor=west, text=black!65] at ($(latchpost.east)+(1.5mm,0)$)
    {Latch post};

\begin{scope}[rotate around={\openAngle:(H)}]
    \draw[panel] (0, -0.5*\panelT) rectangle (\panelW, 0.5*\panelT);
\end{scope}

\draw[sweep] (H) ++(8:\panelW) arc[start angle=8, end angle=82, radius=\panelW];

\coordinate (Cap) at ($(H)+(22.5:\outerR)$);
\draw[searchcap] (Cap) circle (0.27);
\node[font=\small, anchor=west, text=purple!60!black] at ($(Cap)+(0.32,0)$)
    {Search capsule};

\coordinate (DetectCap) at ($(H)+(0:\outerR)$);
\draw[detectcap] (DetectCap) circle (0.30);
\foreach \dx/\dy in {0/0, 1.4/0.9, -1.1/1.2, 1.6/-0.8, -1.4/-0.6, 0.7/1.4, -0.6/-1.1} {
    \fill[green!45!black] (DetectCap) ++(\dx mm, \dy mm) circle (0.6pt);
}

\draw[sweeparrow]
    (H) ++(\openAngle:\outerR) arc[start angle=\openAngle, end angle=8, radius=\outerR];
\draw[sweeparrow]
    (H) ++(\openAngle:\outerR) arc[start angle=\openAngle, end angle=82, radius=\outerR];

\end{tikzpicture}}
    \caption[Latch-side frame post search.]{%
        Latch-side frame post search (overhead view).
        The door panel is shown ajar at $45^\circ$ between hinge and latch frame posts.
        The latch-side edge sweeps along the dotted arc; curved arrows indicate search in either direction along but outside that arc.
        At each candidate angle a vertical search capsule (circle, top-down) counts contained depth points to detect the latch-side post.}
    \label{fig:DoorFrameObjectAlgorithm}
\end{figure}

The estimator places a nominal hinge location at a fixed offset from the mechanism along the panel width.
\autoref{fig:DoorFrameObjectAlgorithm} illustrates our algorithm, which sweeps a vertical 3D capsule around a frame existence hypothesis in the point cloud and, at each candidate angle, uses a CUDA~\cite{nvidia_cuda} points-in-shape counter to search for the latch-side frame post.
As soon as a tunable sufficient number of points are detected within the capsule, that angle is used to define the door frame plane.
Hinge side is inferred from the panel frame relative to the robot viewpoint.
When the panel is clearly open, the sign of the opening angle relative to the hinge side determines push or pull side.

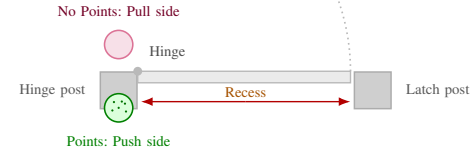
\begin{figure}[h]
\centering
    \resizebox{0.7\columnwidth}{!}{\begin{tikzpicture}[
    font=\small,
    >=Latex,
    post/.style={fill=gray!38, draw=gray!65, line width=0.8pt},
    panel/.style={fill=gray!16, draw=gray!55, line width=0.75pt},
    recessarrow/.style={->, line width=0.75pt, red!70!black},
    checkcap/.style={draw=purple!55, fill=purple!12, line width=0.85pt},
    hingecap/.style={draw=green!50!black, fill=green!14, line width=0.85pt},
    openarc/.style={dotted, line width=0.9pt, gray!55}
]

\def\capR{0.27}

\def\panelW{4.0}
\def\panelT{0.22}
\def\postSize{7mm}
\def\postGap{0.06}
\def\recessH{0.12}
\def\openAngle{20}

\node[post, minimum size=\postSize, inner sep=0pt, anchor=south east] (hingepost)
    at (-\postGap, 0.42-7mm) {};
\node[post, minimum size=\postSize, inner sep=0pt, anchor=south west] (latchpost)
    at (\panelW+\postGap, 0.42-7mm) {};

\coordinate (InnerL) at ($(hingepost.north east)+(\postGap,0)$);
\coordinate (InnerR) at ($(latchpost.north west)+(-\postGap,0)$);

\node[font=\footnotesize, anchor=east, text=black!65] at ($(hingepost.west)+(-1.5mm,0)$)
    {Hinge post};
\node[font=\footnotesize, anchor=west, text=black!65] at ($(latchpost.east)+(1.5mm,0)$)
    {Latch post};

\coordinate (UpperCap) at ($(hingepost.center |- hingepost.north)+(0,{\capR+0.06+2*\capR/3})$);
\coordinate (LowerCap) at ($(hingepost.center |- hingepost.south)+(0,{-\capR+0.12+2*\capR/3})$);
\draw[checkcap] (UpperCap) circle (\capR);
\draw[hingecap] (LowerCap) circle (\capR);
\foreach \dx/\dy in {0/0, 1.2/0.8, -0.9/1.1, 1.4/-0.7, -1.2/-0.5, 0.6/1.2} {
    \fill[green!45!black] (LowerCap) ++(\dx mm, \dy mm) circle (0.6pt);
}
\node[font=\footnotesize, anchor=south, text=purple!60!black]
    at ($(UpperCap)+(0,\capR+4mm)$) {No Points: Pull side};
\node[font=\footnotesize, anchor=north, text=green!50!black]
    at ($(LowerCap)+(0,-\capR-4mm)$) {Points: Push side};

\coordinate (Hinge) at (hingepost.north east);
\coordinate (PanelBotR) at ($(InnerR |- Hinge)+(0,-\panelT)$);
\draw[panel] (Hinge) rectangle (PanelBotR);

\coordinate (LatchTop) at (InnerR |- Hinge);
\draw[openarc]
    let \p1=(Hinge), \p2=(LatchTop),
        \n1={veclen(\x2-\x1,\y2-\y1)},
        \n2={atan2(\y2-\y1,\x2-\x1)} in
    (\p1) ++(\n2:\n1) arc[start angle=\n2, end angle=\n2+\openAngle, radius=\n1];

\coordinate (RecessMid) at ($(InnerL)!0.5!(InnerR) + (0,{-0.04-0.5*\recessH-3*\panelT+1.5*\recessH})$);
\node[font=\footnotesize, anchor=south, text=orange!65!black] at ($(RecessMid)+(0,-0.03)$)
    {Recess};

\draw[recessarrow] (RecessMid) -- (InnerL |- RecessMid);
\draw[recessarrow] (RecessMid) -- (InnerR |- RecessMid);

\fill[gray!55] (Hinge) circle (2.4pt);
\node[font=\footnotesize, anchor=south west, text=black!70] at ($(Hinge)+(1mm,1mm)$)
    {Hinge};

\end{tikzpicture}}
    \caption[Closed-door recess check.]{%
        Push-pull side detection on a closed door (overhead view).
        On the push side, the panel is inset within the frame posts.
        On the hinge/pull side, there is no such recess.
        Looking for depth points in key positions relative to the hinge, as shown, can reveal which side of the door the robot is on.}
    \label{fig:DoorFrameRecessCheck}
\end{figure}

When the door is closed, we use a second vertical capsule depth point check to detect the presence of a door panel recess, which indicates a push door when the recess is present and a pull door when it is not, as illustrated in \autoref{fig:DoorFrameRecessCheck}.

\begin{figure}[h]
\centering
    \includegraphics[width=\columnwidth]{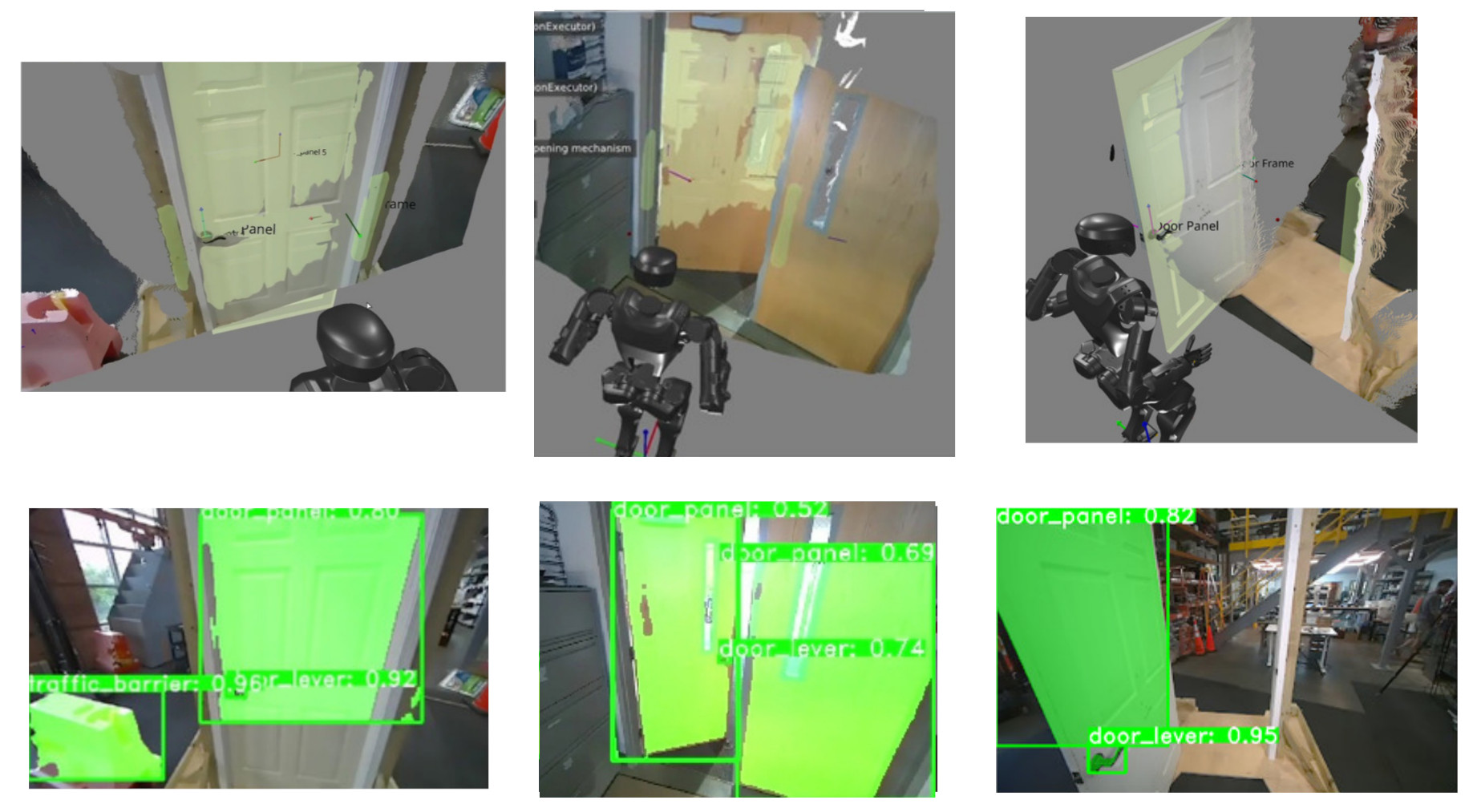}
    \caption[Door state estimation view.]{%
        Left column: A closed lab left pull door. Center column: A real double door, right push, identified as 33$^{\circ}$ ajar. Right column: A right pull door, seen from the panel's backside, detected as being open -81$^{\circ}$.}
    \label{fig:DoorStateEstimation}
\end{figure}

\autoref{fig:DoorStateEstimation} shows the door state estimation examples across closed, ajar, and widely open doors, both hinge configurations, push and pull, and a backside view of the panel.
This information is used for our generalized door behavior to select between type-specific door sub-behaviors.

\subsection{Approach Table Object}
\label{sec:approach_table_object}

\begin{figure}[h]
    \centering
    \resizebox{0.7\columnwidth}{!}{
\newcommand{\TableEdgeFigFont}{\fontsize{9}{10.8}\selectfont}
\newcommand{\TableEdgeFigLabelFont}{\fontsize{8}{9.6}\selectfont}
\newcommand{\TableEdgeFigAxisFont}{\fontsize{7}{8.4}\selectfont}

\begingroup
\fontencoding{OT1}\fontfamily{cmr}\selectfont
\fontsize{10}{12}\selectfont
\begin{tikzpicture}[
    font=\TableEdgeFigFont,
    >=Latex,
    table/.style={fill=gray!20, draw=gray!55, line width=0.75pt},
    sweep/.style={dotted, line width=1pt, gray!55},
    sweeparrow/.style={-{Latex[length=2.0mm,width=1.3mm]}, line width=0.9pt, draw=gray!65},
    searchcap/.style={draw=purple!55, fill=purple!12, line width=0.85pt},
    detectcap/.style={draw=green!50!black, fill=green!14, line width=0.85pt},
    axis/.style={->, line width=0.85pt, gray!70},
    approachaxis/.style={->, line width=0.75pt}
]

\def\capR{0.27}
\def\tableW{5.6}
\def\nearHalfD{0.275}
\def\farHalfD{1.35}
\def\tableAngle{15}
\def\tableCenterY{3.01}
\def\searchForward{0.75}
\def\searchSide{0.87}
\def\approachAxisLen{0.546}
\def\labelClear{6pt}
\def\sweepArrowLen{1.09}

\coordinate (Pelvis) at (0, 0);

\draw[axis] (Pelvis) -- ++(0, \approachAxisLen) node[anchor=south, font=\TableEdgeFigLabelFont] {$X$};
\draw[axis] (Pelvis) -- ++(-\approachAxisLen, 0) node[anchor=east, font=\TableEdgeFigLabelFont] {$Y$};
\node[font=\TableEdgeFigLabelFont, anchor=north, text=black!65] at ($(Pelvis)+(0,-0.18)$)
    {Robot Pelvis};

\coordinate (TableCenter) at (0, \tableCenterY);

\pgfmathsetmacro{\tableAngleRad}{\tableAngle}
\pgfmathsetmacro{\halfW}{0.5*\tableW}
\pgfmathsetmacro{\cosA}{cos(\tableAngleRad)}
\pgfmathsetmacro{\sinA}{sin(\tableAngleRad)}
\coordinate (TableNearL) at ($(TableCenter)+({-\halfW*\cosA+\nearHalfD*\sinA},{-\halfW*\sinA-\nearHalfD*\cosA})$);
\coordinate (TableNearR) at ($(TableCenter)+({\halfW*\cosA+\nearHalfD*\sinA},{\halfW*\sinA-\nearHalfD*\cosA})$);
\path[name path=TableNearEdge, overlay] (TableNearL) -- (TableNearR);

\begin{scope}[rotate around={\tableAngle:(TableCenter)}]
    \draw[table] ($(TableCenter)+(-0.5*\tableW,-\nearHalfD)$)
        rectangle ($(TableCenter)+(0.5*\tableW,\farHalfD)$);
    \node[font=\TableEdgeFigLabelFont, anchor=south, text=black!65]
        at ($(TableCenter)+(0,\farHalfD+0.18)$) {Table};
\end{scope}

\coordinate (StartL) at ($(Pelvis)+(-\searchSide,\searchForward)$);
\coordinate (StartR) at ($(Pelvis)+(\searchSide,\searchForward)$);
\draw[searchcap] (StartL) circle (\capR);
\draw[searchcap] (StartR) circle (\capR);
\node[font=\TableEdgeFigLabelFont, anchor=west, text=purple!60!black]
    at ($(StartR)+(\capR+5mm,0)$) {Search};

\path[name path=SweepL, overlay] (StartL) -- ++(0, 8);
\path[name path=SweepR, overlay] (StartR) -- ++(0, 8);
\path[name intersections={of=SweepL and TableNearEdge, by=DetectL}];
\path[name intersections={of=SweepR and TableNearEdge, by=DetectR}];

\draw[sweep] (StartL) -- (DetectL);
\draw[sweep] (StartR) -- (DetectR);
\draw[sweeparrow] (StartL) -- ++(0, \sweepArrowLen);
\draw[sweeparrow] (StartR) -- ++(0, \sweepArrowLen);

\draw[detectcap] (DetectL) circle (\capR);
\draw[detectcap] (DetectR) circle (\capR);
\node[font=\TableEdgeFigLabelFont, anchor=north west, text=green!50!black]
    at ($(DetectL)+(\capR+\labelClear,-\capR-1.2pt)$) {$c_\mathrm{l}$};
\node[font=\TableEdgeFigLabelFont, anchor=north west, text=green!50!black]
    at ($(DetectR)+(\capR+\labelClear,-\capR-1.2pt)$) {$c_\mathrm{r}$};
\foreach \cap/\dx/\dy in {DetectL/0/0, DetectL/1.4/0.9, DetectL/-1.1/1.2, DetectL/1.6/-0.8, DetectL/-1.4/-0.6, DetectR/0/0, DetectR/1.2/0.8, DetectR/-0.9/1.1, DetectR/1.4/-0.7, DetectR/-1.2/-0.5} {
    \fill[green!45!black] (\cap) ++(\dx mm, \dy mm) circle (0.6pt);
}

\coordinate (ApproachOrigin) at ($(DetectL)!0.5!(DetectR)$);
\begin{scope}[shift={(ApproachOrigin)}, rotate=\tableAngle]
    \node[font=\TableEdgeFigLabelFont, anchor=south, text=black!65, yshift=12pt] at (0, 0.22) {Edge Frame};
    \draw[approachaxis, red!70!black] (0, 0) -- (0, \approachAxisLen)
        node[font=\TableEdgeFigAxisFont, anchor=west, text=red!70!black] {$X$};
    \draw[approachaxis, green!50!black] (0, 0) -- (-\approachAxisLen, 0)
        node[font=\TableEdgeFigAxisFont, midway, above=2pt, xshift=-3.85pt, text=green!50!black] {$Y$};
\end{scope}

\path[use as bounding box]
    ([xshift=-4mm,yshift=-3mm]current bounding box.south west)
    rectangle
    ([xshift=4mm,yshift=3mm]current bounding box.north east);

\end{tikzpicture}
\endgroup}
    \caption[Table edge detection search.]{%
        An overhead 2D view of the table edge detection algorithm.
        Two vertical search capsules start left and right of the robot pelvis and sweep forward along $+X$ until each contains enough depth points on the nearest table edge at $c_\mathrm{l}$ and $c_\mathrm{r}$.
        The red and green axes show the computed approach frame.}
    \label{fig:TableEdgeDetectionAlgorithm}
\end{figure}
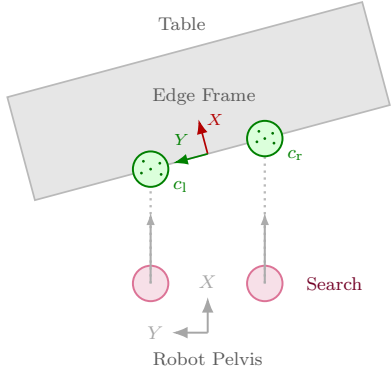

\begin{figure}[h]
    \centering
    \includegraphics[width=\columnwidth]{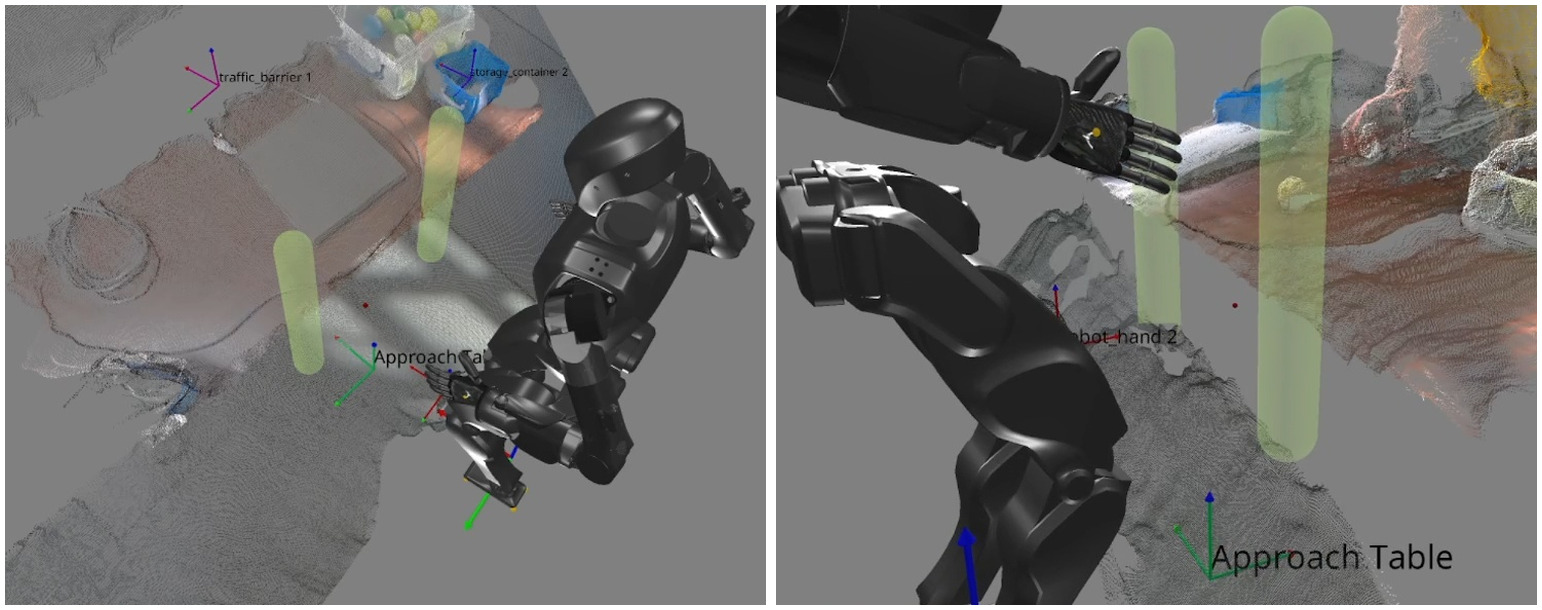}
    \caption[Table approach algorithm visualization.]{%
        The table approach visualization in 3D, where the reference frame available to the physical actions is seen on the floor between the capsules.}
    \label{fig:TableApproachAlgorithm}
\end{figure}

We also developed a novel algorithm for approaching tables accurately and built it into a heuristic scene object called ``approach table''.
This object's algorithm uses only the point cloud and the robot state without relying on semantic detection.
The algorithm detects the table edge and projects a squared-up reference frame onto the ground at the robot's feet.
This is an important capability for doing work on tables because it enables the robot to get close enough to enable arm reachability on the table's workspace without colliding with the table.
The algorithm is illustrated in \autoref{fig:TableEdgeDetectionAlgorithm} and 3D examples are shown on real data in \autoref{fig:TableApproachAlgorithm}.

Two vertical capsules, one on each side of the robot, sweep forward from the pelvis in the mid-feet-under-pelvis frame at a height where table edges are expected.
The capsules start near knee height and end just below chest height so that tables of different heights can be handled with one parameterization.
Our CUDA point counter measures depth point containment inside each capsule at every step.
Each side continues sweeping until enough points are found or the search limit is reached.
The point threshold and search limits are editable in the scene-action settings.
Once both sides have located the edge, the midpoint defines the translational origin and the left-right vector defines the edge orientation.
A pose is projected from this origin onto the ground using the robot's current mid-feet $z$ height.
This enables table approaches that are invariant to table height.

\subsection{Composite and Hybrid Frames}

We define a composite frame as a perception-less object derived from two existing scene frames.
This feature gives the behavior author flexibility by providing a way for them to define behaviors using geometric derivatives of existing scene objects.
Like other privileged objects, composite frames are created by a setup-object scene action and can be referenced by walk, arm, spine, and other physical actions in the same way as a detected object frame.
A representative door setup sequence includes: clear scene, setup door panel, setup door frame, and setup hybrid frame, with the resulting privileged objects visible in the scene panel.

There are currently two composite frame subtypes: an approach frame and a hybrid frame.
Both are parameterized by two source frames by name.

The approach frame orients its $x$ axis to face from frame~A toward frame~B and places its origin on the line segment between them at a tunable distance from frame~B.
This makes it suitable for walking toward a task while stopping short of colliding with the object being approached.
It also resolves approaching objects directly, determining orientation via relative positions, rather than an object-orientation-centric approach that would vary with the object's orientation.

A hybrid frame takes its position from frame~A and its orientation from frame~B.
It is useful for approaching an ajar door panel, where the robot should achieve a stance with the orientation of the door frame but approach the position of the opening mechanism.

Composite frames can also be layered.
For example, an ajar-door hybrid frame can serve as one input to a subsequent approach frame that defines a distant handle approach stance.
Our generalized door traversal behavior constructs panel, frame, hybrid frame, and approach frame objects in this way before selecting a door-specific subtree.

\section{Experimental Evaluation}
\label{ch:defense}

In this section, we present real robot demonstrations organized around speed, resilience, and adaptability.
These demonstrations include six primary task variants: right pull lever door, right push lever door, left pull lever door, picking up and carrying a bottle through a door, sorting colored balls on a table, and sorting colored balls between two tables.
Unless noted otherwise, the evaluations below were run on Alex using onboard stereo color vision and YOLO-based centroid estimation, with no external tracking or fiducial markers.

\subsection{Capability and Speed}

\begin{figure*}[t]
\centering
    {\singlespacing
    \resizebox{\textwidth}{!}{%
        \input{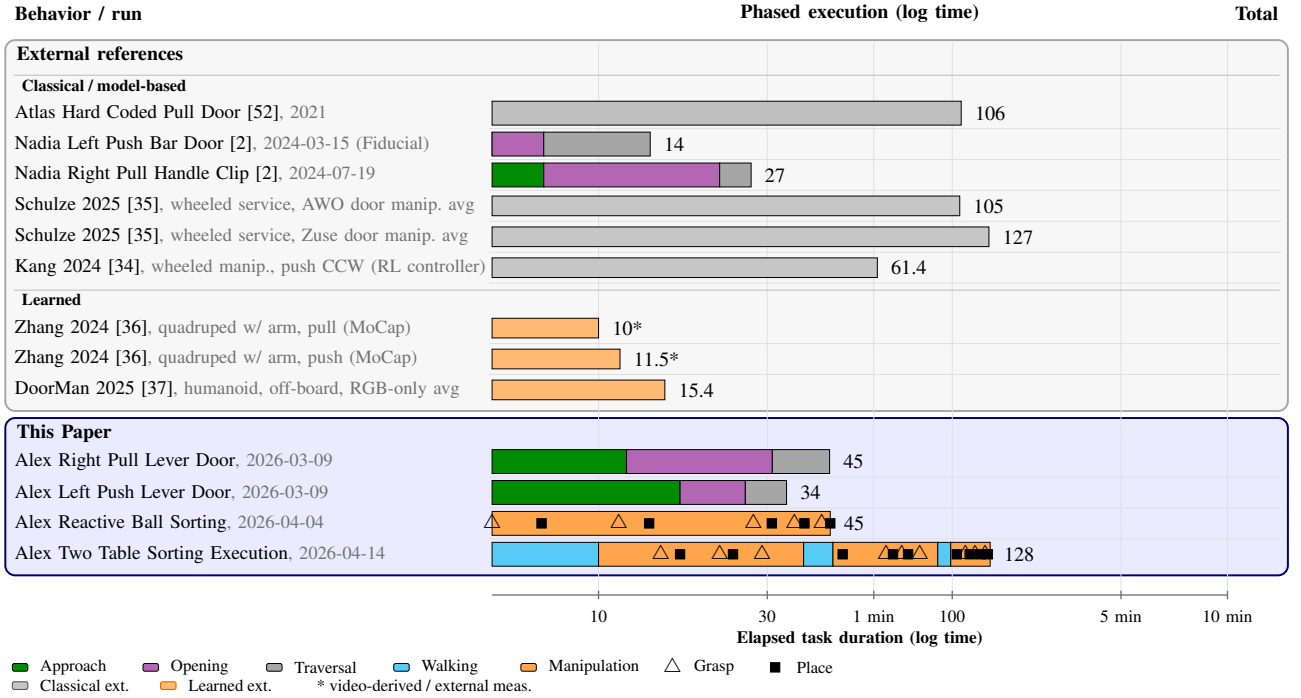}%
    }
    \par
    }
    \caption{
        Real robot task durations on a log-second axis.
        Internal behavior phases are colored for behaviors from this work and for door traversals reported in the literature.
        For door behaviors, it shows approach, opening, and traversal phases, and for loco-manipulation behaviors it shows walking and manipulation phases.
        Triangle and square markers indicate object grasps and placements.
    }
    \label{fig:in_house_speed_phase_timeline}
\end{figure*}

\autoref{fig:in_house_speed_phase_timeline} places the timed loco-manipulation and door behaviors from this work alongside selected door traversals from the literature.
For context, hard-coded Atlas door behaviors from 2021 took over a minute.
Our prior work~\cite{calvert2024behavior} reports drastically faster Nadia door behaviors with runtime-editable structure and concurrent action layering.
In this work, we evaluate more complex and reactive door traversals, tabletop ball sorting, and bottle pickup with door carry-through on Alex.

As literature anchors, the hard-coded Atlas pull-door traversal reported for the June 2021 building-exploration era took 106~s from approach start to full traversal completion, with long pauses, fiducial dependence, and no robot-local authored recovery.
Compared with the July 2024 Nadia right pull handle traversal at 27~s reported in~\cite{calvert2024behavior}, that Atlas run is about 3.9 times slower on a pull-side door task.
The comparison is not controlled, but it situates later runtime-editable door behaviors relative to that early baseline.
The fastest door traversal reported in~\cite{calvert2024behavior} cleared the doorway at 14~s on a push-bar door with continuous walking and concurrent arm actions.

\begin{figure*}[t]
    \centering
        \includegraphics[width=2.0\columnwidth]{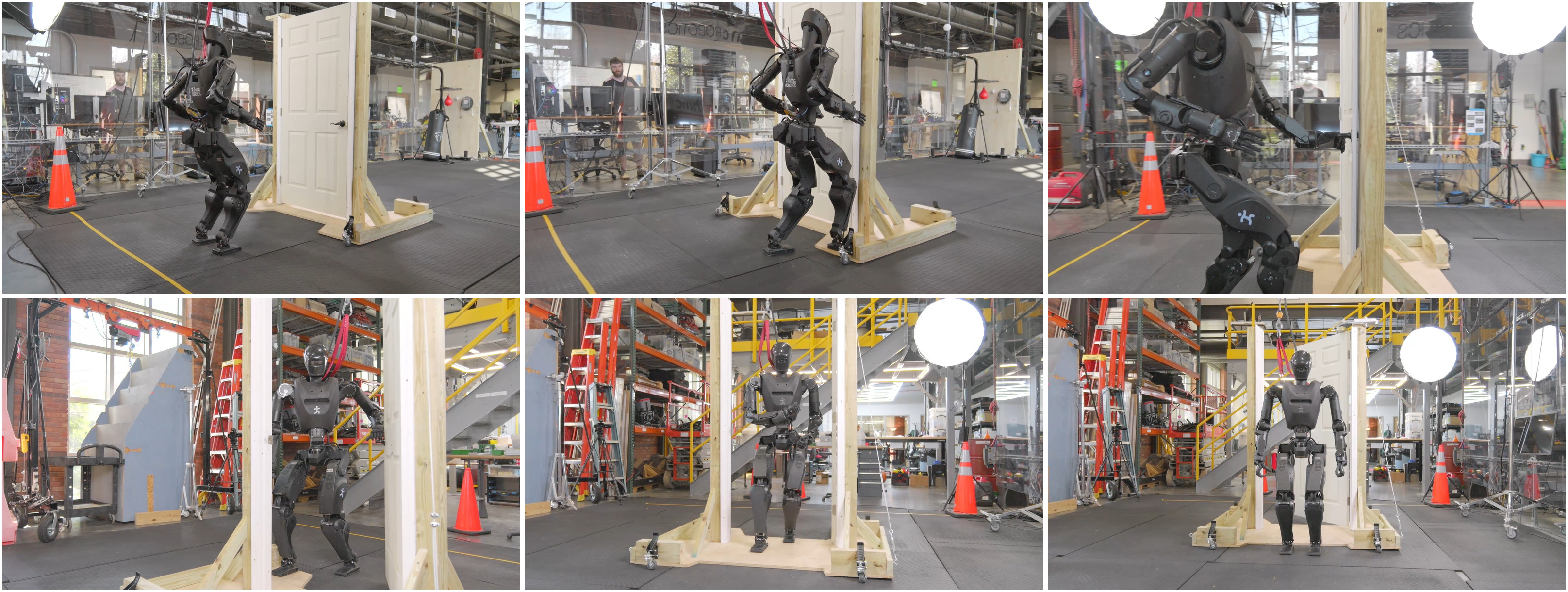}
    \caption[Current Alex right pull lever-handle traversal key frames.]{Current Alex right pull lever-handle traversal key frames.
    From left to right and top to bottom: coarse approach begins at 0:02, fine approach begins at 0:09, opening stance is achieved at 0:14, the door is unlatched at 0:18, the door is opened and traversal begins at 0:33, the shoulders align with the frame at 0:42, and traversal completes at 0:47. Supplementary video material accompanies this submission.}
    \label{fig:AlexRightPullDoorProfessional}
\end{figure*}

On Alex, we measured a right pull lever-handle door traversal in 45 seconds, shown in \autoref{fig:AlexRightPullDoorProfessional}.
This run comprises a 12~s approach, 4~s unlatch, 15~s panel opening, and 14~s traversal as summarized in \autoref{tab:alex_right_pull_phase_breakdown}.
The door panel opening motion takes time on pull doors because the robot must stay clear of the swing.
One notable limiter on speed compared to our prior work was the limited spine yaw range of motion on this version of Alex, +/- $30^\circ$, instead of the +/- $60^\circ$ available on Nadia.
This limitation meant that more intricate motions were required to open the door---it could not be done in one sweeping motion.
However, this timing still places Alex pull traversals in the same tens of seconds regime as the 2024 Nadia results reported in~\cite{calvert2024behavior} and well below the minute-scale classical literature entries in \autoref{fig:in_house_speed_phase_timeline}.

\begin{table}[h]
\caption{Alex right pull lever-handle traversal timing.}
\centering
\setlength{\tabcolsep}{3pt}
\renewcommand{\arraystretch}{1.01}
{\small
\begin{tabular}{@{}c c >{\raggedright\arraybackslash}p{0.56\columnwidth}@{}}
 \hline
 Time window & Duration & Description \\
 \hline
 0:02--0:14 & 12~s & Coarse approach from the distant start, followed by the fine approach into the authored opening stance. \\
 0:14--0:18 & 4~s & Final upper-body alignment, handle engagement, and latch release. \\
 0:18--0:33 & 15~s & Pull the panel clear while staying outside the door swing. \\
 0:33--0:47 & 14~s & Commit through the frame, bring the shoulders into alignment by 0:42, and clear the doorway. \\
 \hline
\end{tabular}}
\label{tab:alex_right_pull_phase_breakdown}
\end{table}

\subsection{Reliability}

We conducted a repeated-run trial of door approach and opening.
The trial produced 11/11 successful push-door approaches and openings and 12/12 successful pull-door approaches and openings, ending the experiment before failure was reached.
All trials used the same lab door on the same night.
We did not repeat full doorway traversal in this test, because we determined that the walking control was not working well enough and it would have risked damage to the hardware, preventing a test of reliability of the rest of the system.
However, these repeated trials still demonstrate reliable approach and opening.

\subsection{Resilience}
\label{sec:resilience}

To evaluate disturbance recovery, we ran a left pull door traversal with authored fallback nodes and spatial occupancy-checking condition nodes.
A human disturbed the robot four times during the opening phase by pulling the door out of the robot's hand from the other side.
The failure to open the door was detected via an absence of points where the door panel should be after the door opening screw primitive action.
A human also blocked the doorway before traversal and the robot waited until the doorway was clear before proceeding.
Across these five disturbances, the run finished in 65~s overall.
This reactivity was authored at runtime, in contrast to the hard-coded door reactivity demo on Nadia presented in~\cite{calvert2024behavior}.

\begin{figure}[h]
    \centering
    \includegraphics[width=1.0\columnwidth]{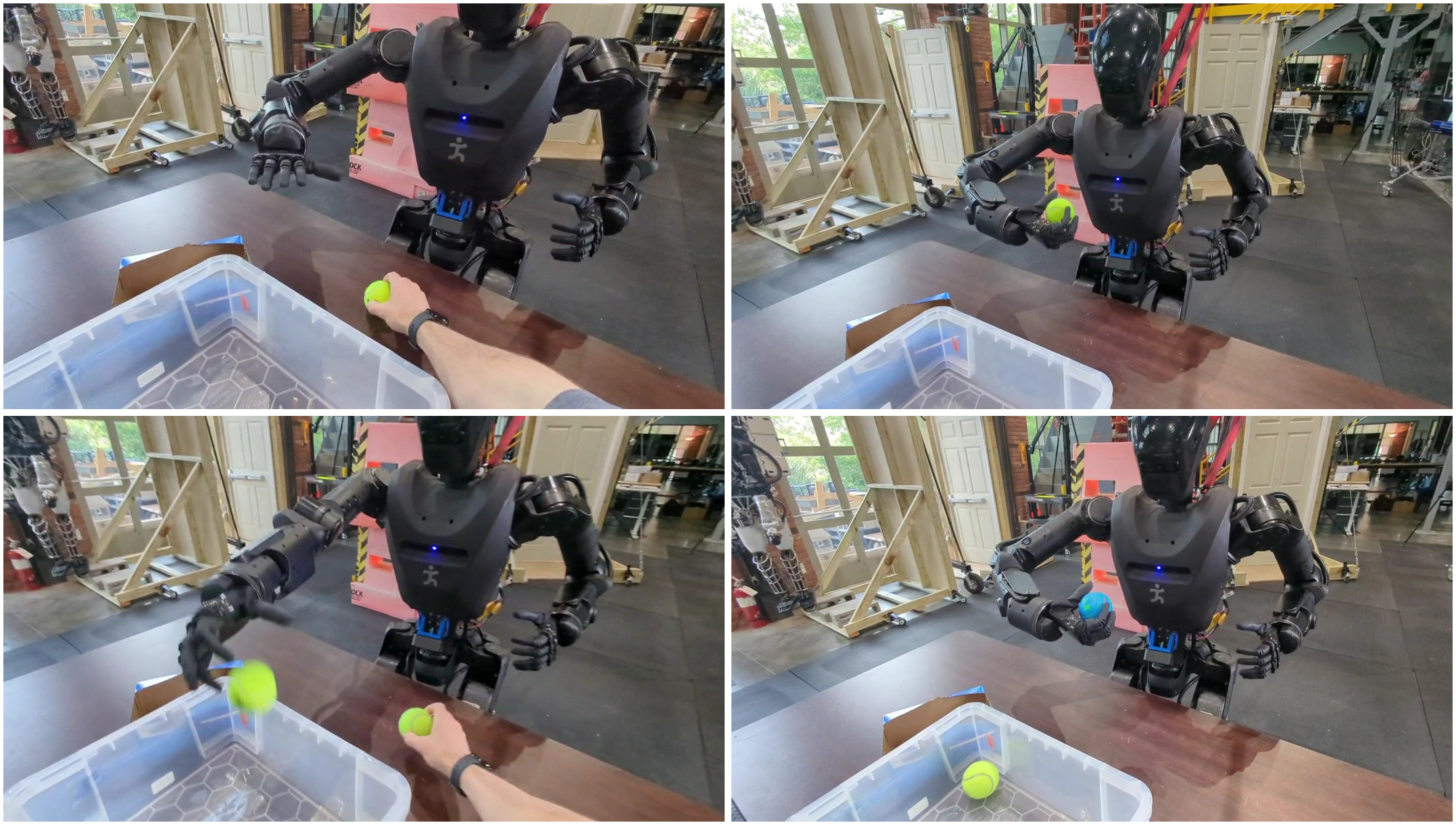}
    \caption[Alex reactive single-table ball sorting.]{A reactive single-table colored ball sorting behavior run.
    The run includes a human disturbance; the tree abandons a stale pick and returns to search when the hand-centered containment check shows that the object is no longer in the hand.
    It correctly sorts balls by color into the bins: 5 yellow and 1 blue.}
    \label{fig:ReactiveRobustBallSortingDisturbance}
\end{figure}

We also evaluated a reactive single-table ball-sorting behavior while humans continuously placed and sometimes removed balls on the table, shown in \autoref{fig:ReactiveRobustBallSortingDisturbance}.
The run completed in 45.2 seconds.
The behavior detected a failed pick at 17.8~s and aborted the place, returning to the search phase.
The robot still correctly sorted all six of the balls that were not removed at the last second, placing the blue ball in the blue container.
This rate of action supports hypothesis 1 on manipulation speed under a dynamic scene.

\subsection{Fast Behavior Authoring}
\label{sec:h2_fast_behavior_authoring}

\begin{figure*}[t]
\centering
    {\singlespacing
    \resizebox{\textwidth}{!}{%
        \newcommand{\inauthlabel}[2]{#1{\fontsize{7.5}{7.5}\selectfont\color{black!55}{, #2}}}
\newcommand{\inauthrowlabel}[3]{%
  \node[rowlbl, anchor=west] at (\systemleft + 0.02, #1) {\inauthlabel{#2}{#3}};}
\newcommand{\inauthnum}[3]{%
  \pgfmathparse{#2 > 9.8 ? 1 : 0}%
  \ifnum\pgfmathresult=1
    \node[numlbl, anchor=east] at (\numright - 0.10, #1) {#3};%
  \else
    \pgfmathsetmacro{\inauthnumx}{\authscopedleft + \authxscale * (#2 - \authlegacyaxisleft) + 0.08 - \authbarshift}%
    \node[numlbl, anchor=west] at (\inauthnumx, #1) {#3};%
  \fi}
\newcommand{\authbarphase}[5]{%
  \filldraw[#5, draw=black, line width=0.28pt]
    (#3, {#1 - #2}) rectangle (#4, {#1 + #2});}
\newcommand{\authbarsuccess}[2]{%
  \node[draw=black, line width=0.28pt, fill=green!55!black, rounded corners=0.6pt, inner xsep=0.8pt, inner ysep=0.4pt, font=\fontsize{6.0}{6.4}\selectfont\bfseries, text=white] at (#2, #1) {$\checkmark$};}
\newcommand{\inauthswatch}[3]{%
  \filldraw[#1, rounded corners=0.8pt] (#2, {\inlegendy - 0.055}) rectangle ++(0.22, 0.11);%
  \node[legendtxt, anchor=west] at ({#2 + 0.28}, \inlegendy) {#3};}

\begin{tikzpicture}[
    x=1cm,
    y=1cm,
    every node/.style={outer sep=0pt},
    rowlbl/.style={font=\fontsize{7.8}{8.4}\selectfont, anchor=west, align=left, text width=\labeltextw cm},
    numlbl/.style={font=\fontsize{8.2}{8.8}\selectfont, anchor=west, align=left, text width=0.82cm},
    bandtitle/.style={font=\footnotesize\bfseries, anchor=north west, inner xsep=1.2pt, inner ysep=0.8pt},
    legendtxt/.style={font=\fontsize{6.8}{7.4}\selectfont}
  ]
  \def\barh{0.17}

  \def\systemw{6.85}
  \def\barw{10.55}
  \def\numw{0.78}
  \def\labeltextw{6.77}
  \def\colgap{0.04}
  \def\headerh{0.95}
  \def\rowh{0.44}
  \def\bandgap{0.10}
  \def\bandtitleh{0.40}
  \def\bandpadbottom{0.08}
  \def\bottomregionh{2.02}

  \def\authlegacyaxisleft{4.01}
  \def\authlegacyaxisright{10.4028}

  \pgfmathsetmacro{\systemleft}{0}
  \pgfmathsetmacro{\systemright}{\systemleft + \systemw}
  \pgfmathsetmacro{\barleft}{\systemright + \colgap}
  \pgfmathsetmacro{\barright}{\barleft + \barw}
  \pgfmathsetmacro{\numleft}{\barright + \colgap}
  \pgfmathsetmacro{\numright}{\numleft + \numw}
  \pgfmathsetmacro{\figurewidth}{\numright}

  \pgfmathsetmacro{\bandah}{\bandtitleh + \rowh + \bandpadbottom}
  \pgfmathsetmacro{\bandbh}{\bandtitleh + 3 * \rowh + \bandpadbottom}
  \pgfmathsetmacro{\figureheight}{\headerh + \bandah + \bandgap + \bandbh + \bottomregionh}

  \pgfmathsetmacro{\headertop}{\figureheight}
  \pgfmathsetmacro{\headerbottom}{\headertop - \headerh}
  \pgfmathsetmacro{\bandatop}{\headerbottom}
  \pgfmathsetmacro{\bandabottom}{\bandatop - \bandah}
  \pgfmathsetmacro{\bandbtop}{\bandabottom - \bandgap}
  \pgfmathsetmacro{\bandbbottom}{\bandbtop - \bandbh}

  \pgfmathsetmacro{\rownadia}{\bandatop - \bandtitleh - 0.5 * \rowh}
  \pgfmathsetmacro{\rowunitree}{\bandbtop - \bandtitleh - 0.5 * \rowh}
  \pgfmathsetmacro{\rowalexpull}{\rowunitree - \rowh}
  \pgfmathsetmacro{\rowalexpush}{\rowalexpull - \rowh}

  \def\authbarshift{0.30}

  \pgfmathsetmacro{\authaxisleft}{\barleft + 0.04}
  \pgfmathsetmacro{\authaxisright}{\barright - 0.04}
  \pgfmathsetmacro{\authscopedleft}{\authaxisleft - \authbarshift}
  \pgfmathsetmacro{\authscopedright}{\authaxisright - \authbarshift}
  \pgfmathsetmacro{\authxscale}{(\authscopedright - \authscopedleft)/(\authlegacyaxisright - \authlegacyaxisleft)}

  \pgfmathsetmacro{\axisy}{\bandbbottom - 0.28}
  \def\legendentryh{0.10}
  \def\legendrowpitch{0.26}
  \pgfmathsetmacro{\axistitley}{\axisy - 0.40}
  \pgfmathsetmacro{\axistitlebottom}{\axistitley - 0.14}
  \pgfmathsetmacro{\legendgapbelowtitle}{0.44}
  \pgfmathsetmacro{\inlegendrowA}{\axistitlebottom - \legendgapbelowtitle - 0.5 * \legendentryh}
  \pgfmathsetmacro{\inlegendrowB}{\inlegendrowA - \legendrowpitch}
  \pgfmathsetmacro{\durationheaderx}{\authaxisleft - 0.28}

  \def\legenditemsep{0.36}
  \def\legendleft{0.10}
  \pgfmathsetmacro{\legAuthAone}{\legendleft}
  \pgfmathsetmacro{\legAuthAtwo}{\legAuthAone + 2.05 + \legenditemsep}
  \pgfmathsetmacro{\legAuthAthree}{\legAuthAtwo + 2.45 + \legenditemsep}
  \pgfmathsetmacro{\legAuthAfour}{\legAuthAthree + 2.45 + \legenditemsep}
  \pgfmathsetmacro{\legAuthAfive}{\legAuthAfour + 2.25 + \legenditemsep}
  \pgfmathsetmacro{\legAuthBone}{\legendleft}
  \pgfmathsetmacro{\legAuthBtwo}{\legAuthBone + 3.36 + \legenditemsep}

  \path[draw=black!35, fill=gray!6, rounded corners=4pt, line width=0.8pt]
    (0, \bandabottom) rectangle (\figurewidth, \bandatop);
  \path[draw=green!45!black, fill=green!10, rounded corners=4pt, line width=0.8pt]
    (0, \bandbbottom) rectangle (\figurewidth, \bandbtop);

  \node[anchor=west, font=\footnotesize\bfseries] at (\systemleft + 0.02, \headerbottom + 0.38) {Behavior / session};
  \node[anchor=east, font=\footnotesize\bfseries] at (\numright - 0.02, \headerbottom + 0.38) {Total};

  \inauthrowlabel{\rownadia}{Nadia Scratch Right Push Opening~\cite{calvert2023authoring}}{2023}
  \inauthrowlabel{\rowunitree}{Unitree H1-2 Right Pull Opening Loop}{2026-01-02}
  \inauthrowlabel{\rowalexpull}{Alex First Right Pull Door Bring Up}{2026-01-20--24}
  \inauthrowlabel{\rowalexpush}{Alex Scratch Left Push Door}{2026-02-22}

  \begin{scope}[shift={(-\authbarshift,0)}]
  \node[anchor=west, font=\footnotesize\bfseries] at (\durationheaderx, \headerbottom + 0.38) {Scratch authoring duration (log time)};
  \begin{scope}[shift={(\authscopedleft,0)}, xscale=\authxscale, shift={(-\authlegacyaxisleft,0)}]

  \foreach \tickx in {5.8974,6.8801,7.8627,8.8453,9.8280,10.4028} {
    \draw[draw=black!12, line width=0.4pt] (\tickx, \axisy) -- (\tickx, \bandatop - 0.04);
  }

  \authbarphase{\rownadia}{\barh}{\authlegacyaxisleft}{4.8168}{blue!45}
  \authbarphase{\rownadia}{\barh}{4.8168}{5.6945}{violet!60}
  \authbarphase{\rownadia}{\barh}{5.6945}{6.0325}{black!35}

  \authbarphase{\rowunitree}{\barh}{\authlegacyaxisleft}{5.7732}{blue!45}
  \authbarphase{\rowunitree}{\barh}{5.7732}{5.9643}{violet!60}
  \authbarphase{\rowunitree}{\barh}{5.9643}{5.9763}{black!35}
  \authbarsuccess{\rowunitree}{5.9763}

  \authbarphase{\rowalexpull}{\barh}{\authlegacyaxisleft}{6.0318}{blue!45}
  \authbarphase{\rowalexpull}{\barh}{6.0318}{6.9455}{green!55!black}
  \authbarphase{\rowalexpull}{\barh}{6.9455}{7.4222}{green!55!black}
  \authbarphase{\rowalexpull}{\barh}{7.4222}{8.5552}{violet!60}
  \authbarphase{\rowalexpull}{\barh}{8.5552}{8.8280}{violet!60}
  \authbarphase{\rowalexpull}{\barh}{8.8280}{9.4755}{violet!60}
  \authbarphase{\rowalexpull}{\barh}{9.4755}{9.8070}{violet!60}
  \authbarphase{\rowalexpull}{\barh}{9.8070}{10.1271}{black!35}
  \authbarphase{\rowalexpull}{\barh}{10.1271}{10.2667}{orange!70}
  \authbarphase{\rowalexpull}{\barh}{10.2667}{10.2836}{orange!70}
  \authbarphase{\rowalexpull}{\barh}{10.2836}{10.3013}{black!35}
  \authbarsuccess{\rowalexpull}{10.3013}

  \authbarphase{\rowalexpush}{\barh}{\authlegacyaxisleft}{5.6026}{blue!45}
  \authbarphase{\rowalexpush}{\barh}{5.6026}{7.5950}{violet!60}
  \authbarphase{\rowalexpush}{\barh}{7.5950}{7.7316}{black!35}
  \authbarphase{\rowalexpush}{\barh}{7.7316}{7.8419}{orange!70}
  \authbarphase{\rowalexpush}{\barh}{7.8419}{7.8603}{black!35}
  \authbarsuccess{\rowalexpush}{7.8603}

  \draw[draw=black!55, line width=0.55pt] (\authlegacyaxisleft, \axisy) -- (\authlegacyaxisright, \axisy);
  \foreach \tickx/\ticklabel in {5.8974/30 min,6.8801/1 h,7.8627/2 h,8.8453/4 h,9.8280/8 h,10.4028/12 h} {
    \draw[draw=black!55, line width=0.55pt] (\tickx, \axisy - 0.06) -- (\tickx, \axisy + 0.06);
    \node[font=\scriptsize, anchor=north] at (\tickx, \axisy - 0.10) {\ticklabel};
  }
  \node[font=\scriptsize\bfseries, anchor=north] at ({0.5 * (\authlegacyaxisleft + \authlegacyaxisright)}, \axistitley) {Normalized active authoring duration (log time)};
  \end{scope}
  \end{scope}

  \inauthnum{\rownadia}{6.0325}{33}
  \inauthnum{\rowunitree}{5.9763}{31:43}
  \inauthnum{\rowalexpull}{10.3013}{11:10:17}
  \inauthnum{\rowalexpush}{7.8603}{1:59:48}

  \foreach \rowy in {\rownadia,\rowunitree,\rowalexpull} {
    \draw[draw=black!12, line width=0.4pt] (\barleft, {\rowy - 0.5 * \rowh}) -- (\figurewidth - 0.10, {\rowy - 0.5 * \rowh});
  }

  \node[bandtitle, fill=gray!6] at (0.12, \bandatop - 0.06) {External references};
  \node[bandtitle, fill=green!10] at (0.12, \bandbtop - 0.06) {This Paper};

  \pgfmathsetmacro{\inlegendy}{\inlegendrowA}
  \inauthswatch{blue!45, draw=black, line width=0.28pt}{\legAuthAone}{Setup / structure}
  \inauthswatch{green!55!black, draw=black, line width=0.28pt}{\legAuthAtwo}{Approach / locomotion}
  \inauthswatch{violet!60, draw=black, line width=0.28pt}{\legAuthAthree}{Opening / manipulation}
  \inauthswatch{black!35, draw=black, line width=0.28pt}{\legAuthAfour}{Traversal / completion}
  \inauthswatch{orange!70, draw=black, line width=0.28pt}{\legAuthAfive}{Fix / recovery}

  \pgfmathsetmacro{\inlegendy}{\inlegendrowB}
  \node[draw=black, line width=0.28pt, fill=green!55!black, rounded corners=0.6pt, inner xsep=0.8pt, inner ysep=0.4pt, font=\fontsize{6.0}{6.4}\selectfont\bfseries, text=white] at (\legAuthBone + 0.11, \inlegendy) {$\checkmark$};
  \node[legendtxt, anchor=west] at ({\legAuthBone + 0.28}, \inlegendy) {First autonomous success};
\end{tikzpicture}%
    }
    \par
    }
    \caption{
        In-house real-robot from-scratch authoring durations on a log-second axis with internal milestone structure.
        The External references band shows a prior Nadia stand-in-place opening with ArUco perception and teleoperated traversal walk~\cite{calvert2023authoring}.
        The band for our paper groups Unitree H1-2 and Alex sessions from this work.
        Checkmarks signify the first autonomous success.
        Gaps between authoring sessions are excluded for the multi-day Alex bring-up.
    }
    \label{fig:in_house_authoring_from_scratch_timeline}
\end{figure*}
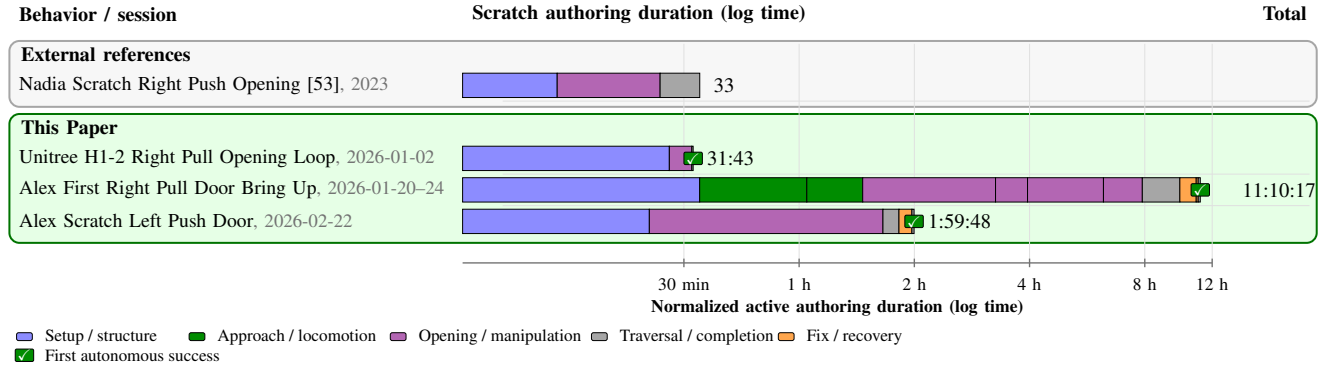

We measured from-scratch authoring sessions on real hardware, with active authoring time recovered from screen recordings.
\autoref{fig:in_house_authoring_from_scratch_timeline} summarizes these durations on a log-second axis with internal milestone structure.
Sessions were performed by an expert operator using the runtime-editable tree and scene interface described in \autoref{ch:architecture2}.

A prior model-based runtime-editable implementation reached a measured 33 minutes of active authoring for a stand-in-place opening on Nadia~\cite{calvert2023authoring}, but the traversal walk-through was teleoperated and perception used a fiducial marker, so that row is not directly comparable to the fully autonomous, marker-free sessions below.

On the Unitree H1-2, we authored a looping standing right pull lever-handle opening behavior from an empty behavior tree.
The session reached first autonomous looping execution in 31 minutes and 43 seconds of active authoring time.
The goto loop added at 31:27 immediately began the 32/32 repeated opening test.
This session shows the diagnose, edit, and retest loop in one sitting, which is what hypothesis 2 claims relative to redeploy or retrain workflows.

\begin{figure*}[t]
    \centering
    \includegraphics[width=2.0\columnwidth]{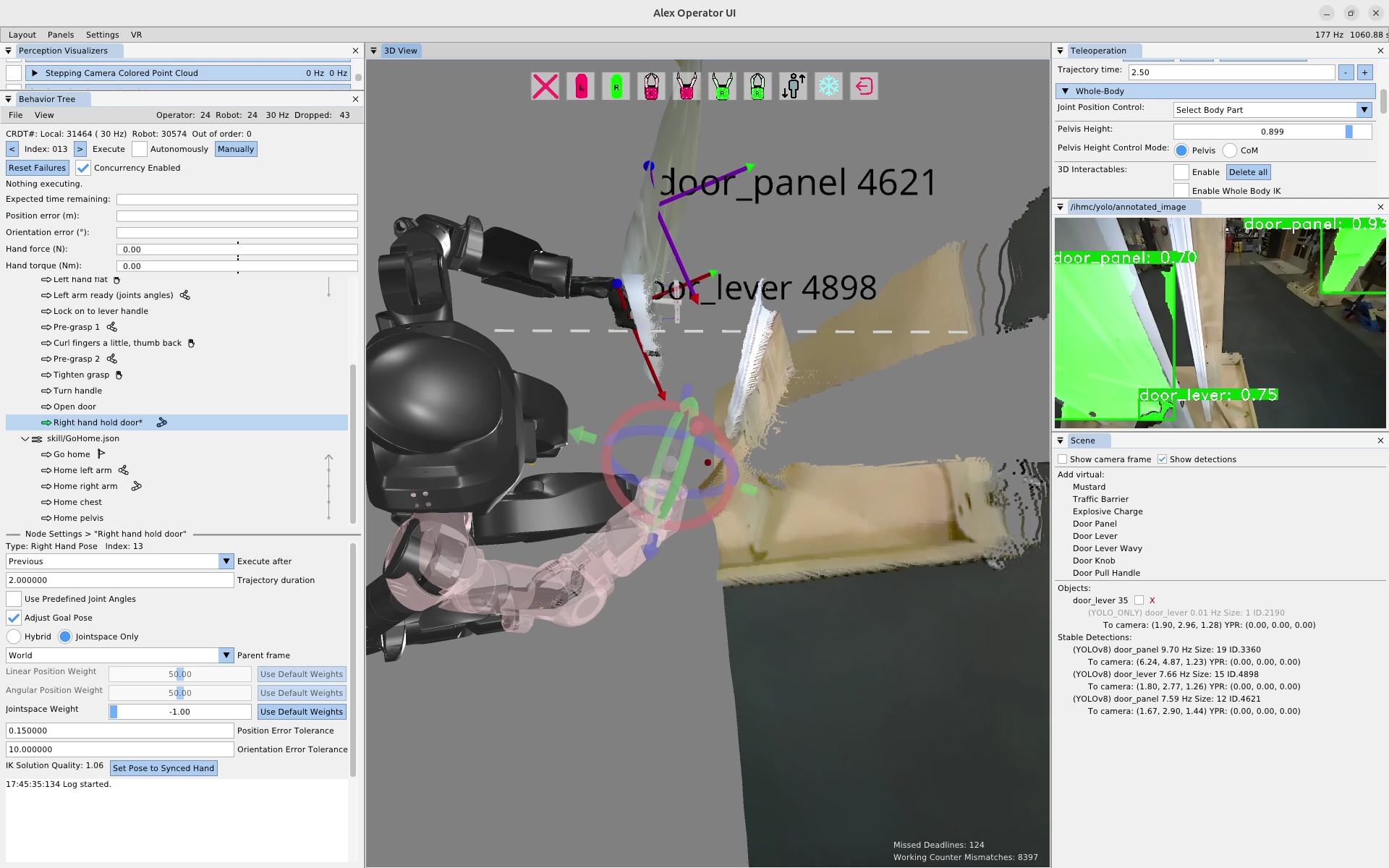}
    \caption{A screenshot of the authoring interface during Alex's first pull door session.
    On the left, the behavior tree can be seen with a ``Right hand hold door'' arm action selected.
    In the bottom left, settings for the selected node are shown.
    In the center, the robot digital twin is rendered in 3D with the color point cloud from the ZED X Mini.
    A pose gizmo is being used to tune the selected arm action's configuration.
    Red and blue coordinate frames ``door\_panel'' and ``door\_lever'' show the current 3D positions of the YOLO detected objects.
    On the right, a first-person view is shown with the YOLO segmentation areas highlighted.
    In the bottom right, the behavior scene state is shown, the privileged objects area and the stable persistent detections below.}
    \label{fig:h2_alex_first_pull_authoring}
\end{figure*}

On Alex, the first full pull lever door traversal brought up from scratch required 11 hours, 10 minutes, and 17 seconds of measured active authoring across 5 sessions over 5 days, excluding inactive gaps, as shown in \autoref{fig:h2_alex_first_pull_authoring}.
Major milestones include the first open from a staggered stance on day 3 and a hold-open fallback added after force-related trouble on day 5.
This long duration shows the cost of bringing up the first door traversal behavior on a new hardware platform.

A subsequent left push door traversal on Alex was authored from an empty sequence to first fully automatic success in 1 hour, 59 minutes, and 48 seconds of active authoring.
A failure at 1:50:11 from an arm taskspace error was fixed in about 9 minutes by bringing the right arm in using jointspace for traversal.

This evidence shows that our behavior system supports authoring novel behaviors in the hours regime.
The very first door behavior on Alex, in which lots of robot-specific things had to be worked out, took only around 11 hours of authoring to complete an autonomous run.
A push door behavior was then authored on Alex within 2 hours.
That short edit and retest loop stands in contrast to multi-day retraining pipelines.

\subsection{Fast Adaptation, Extension, and Combination of Behaviors}
\label{sec:h3_fast_adaptation}

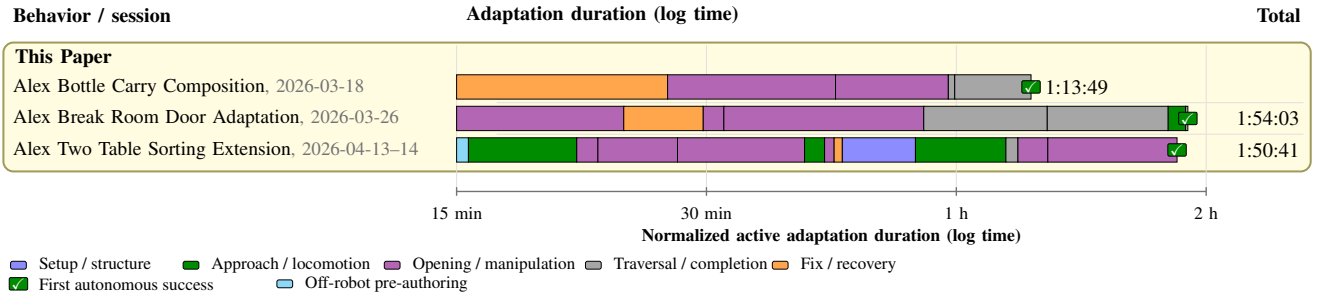
\begin{figure*}[t]
\centering
    {\singlespacing
    \resizebox{\textwidth}{!}{%
        \newcommand{\inadaptlabel}[2]{#1{\fontsize{7.5}{7.5}\selectfont\color{black!55}{, #2}}}
\newcommand{\inadaptrowlabel}[3]{%
  \node[rowlbl, anchor=west] at (\systemleft + 0.02, #1) {\inadaptlabel{#2}{#3}};}
\newcommand{\inadaptnum}[3]{%
  \pgfmathparse{#2 > 9.8 ? 1 : 0}%
  \ifnum\pgfmathresult=1
    \node[numlbl, anchor=east] at (\numright - 0.10, #1) {#3};%
  \else
    \pgfmathsetmacro{\inadaptnumx}{\adaptscopedleft + \adaptxscale * (#2 - \adaptlegacyaxisleft) + 0.08 - \adaptbarshift}%
    \node[numlbl, anchor=west] at (\inadaptnumx, #1) {#3};%
  \fi}
\newcommand{\adaptbarphase}[5]{%
  \filldraw[#5, draw=black, line width=0.28pt]
    (#3, {#1 - #2}) rectangle (#4, {#1 + #2});}
\newcommand{\adaptbarsuccess}[2]{%
  \node[draw=black, line width=0.28pt, fill=green!55!black, rounded corners=0.6pt, inner xsep=0.8pt, inner ysep=0.4pt, font=\fontsize{6.0}{6.4}\selectfont\bfseries, text=white] at (#2, #1) {$\checkmark$};}
\newcommand{\inadaptswatch}[3]{%
  \filldraw[#1, rounded corners=0.8pt] (#2, {\inlegendy - 0.055}) rectangle ++(0.22, 0.11);%
  \node[legendtxt, anchor=west] at ({#2 + 0.28}, \inlegendy) {#3};}

\begin{tikzpicture}[
    x=1cm,
    y=1cm,
    every node/.style={outer sep=0pt},
    rowlbl/.style={font=\fontsize{7.8}{8.4}\selectfont, anchor=west, align=left, text width=\labeltextw cm},
    numlbl/.style={font=\fontsize{8.2}{8.8}\selectfont, anchor=west, align=left, text width=0.82cm},
    bandtitle/.style={font=\footnotesize\bfseries, anchor=north west, inner xsep=1.2pt, inner ysep=0.8pt},
    legendtxt/.style={font=\fontsize{6.8}{7.4}\selectfont}
  ]
  \def\barh{0.17}

  \def\systemw{6.85}
  \def\barw{10.55}
  \def\numw{0.78}
  \def\labeltextw{6.77}
  \def\colgap{0.04}
  \def\headerh{0.95}
  \def\rowh{0.44}
  \def\bandtitleh{0.40}
  \def\bandpadbottom{0.08}
  \def\bottomregionh{2.02}

  \def\adaptlegacyaxisleft{4.3100}
  \def\adaptlegacyaxisright{10.5800}

  \pgfmathsetmacro{\systemleft}{0}
  \pgfmathsetmacro{\systemright}{\systemleft + \systemw}
  \pgfmathsetmacro{\barleft}{\systemright + \colgap}
  \pgfmathsetmacro{\barright}{\barleft + \barw}
  \pgfmathsetmacro{\numleft}{\barright + \colgap}
  \pgfmathsetmacro{\numright}{\numleft + \numw}
  \pgfmathsetmacro{\figurewidth}{\numright}

  \pgfmathsetmacro{\bandch}{\bandtitleh + 3 * \rowh + \bandpadbottom}
  \pgfmathsetmacro{\figureheight}{\headerh + \bandch + \bottomregionh}

  \pgfmathsetmacro{\headertop}{\figureheight}
  \pgfmathsetmacro{\headerbottom}{\headertop - \headerh}
  \pgfmathsetmacro{\bandctop}{\headerbottom}
  \pgfmathsetmacro{\bandcbottom}{\bandctop - \bandch}

  \pgfmathsetmacro{\rowbottle}{\bandctop - \bandtitleh - 0.5 * \rowh}
  \pgfmathsetmacro{\rowbreak}{\rowbottle - \rowh}
  \pgfmathsetmacro{\rowtwotable}{\rowbreak - \rowh}

  \def\adaptbarshift{0.30}

  \pgfmathsetmacro{\adaptaxisleft}{\barleft + 0.04}
  \pgfmathsetmacro{\adaptaxisright}{\barright - 0.04}
  \pgfmathsetmacro{\adaptscopedleft}{\adaptaxisleft - \adaptbarshift}
  \pgfmathsetmacro{\adaptscopedright}{\adaptaxisright - \adaptbarshift}
  \pgfmathsetmacro{\adaptxscale}{(\adaptscopedright - \adaptscopedleft)/(\adaptlegacyaxisright - \adaptlegacyaxisleft)}

  \pgfmathsetmacro{\axisy}{\bandcbottom - 0.28}
  \def\legendentryh{0.10}
  \def\legendrowpitch{0.26}
  \pgfmathsetmacro{\axistitley}{\axisy - 0.40}
  \pgfmathsetmacro{\axistitlebottom}{\axistitley - 0.14}
  \pgfmathsetmacro{\legendgapbelowtitle}{0.44}
  \pgfmathsetmacro{\inlegendrowA}{\axistitlebottom - \legendgapbelowtitle - 0.5 * \legendentryh}
  \pgfmathsetmacro{\inlegendrowB}{\inlegendrowA - \legendrowpitch}
  \pgfmathsetmacro{\durationheaderx}{\adaptaxisleft - 0.28}

  \def\legenditemsep{0.36}
  \def\legendleft{0.10}
  \pgfmathsetmacro{\legAdaptAone}{\legendleft}
  \pgfmathsetmacro{\legAdaptAtwo}{\legAdaptAone + 2.05 + \legenditemsep}
  \pgfmathsetmacro{\legAdaptAthree}{\legAdaptAtwo + 2.45 + \legenditemsep}
  \pgfmathsetmacro{\legAdaptAfour}{\legAdaptAthree + 2.45 + \legenditemsep}
  \pgfmathsetmacro{\legAdaptAfive}{\legAdaptAfour + 2.25 + \legenditemsep}
  \pgfmathsetmacro{\legAdaptBone}{\legendleft}
  \pgfmathsetmacro{\legAdaptBtwo}{\legAdaptBone + 3.36 + \legenditemsep}

  \path[draw=yellow!55!black, fill=yellow!14, rounded corners=4pt, line width=0.8pt]
    (0, \bandcbottom) rectangle (\figurewidth, \bandctop);

  \node[anchor=west, font=\footnotesize\bfseries] at (\systemleft + 0.02, \headerbottom + 0.38) {Behavior / session};
  \node[anchor=east, font=\footnotesize\bfseries] at (\numright - 0.02, \headerbottom + 0.38) {Total};

  \inadaptrowlabel{\rowbottle}{Alex Bottle Carry Composition}{2026-03-18}
  \inadaptrowlabel{\rowbreak}{Alex Break Room Door Adaptation}{2026-03-26}
  \inadaptrowlabel{\rowtwotable}{Alex Two Table Sorting Extension}{2026-04-13--14}

  \begin{scope}[shift={(-\adaptbarshift,0)}]
  \node[anchor=west, font=\footnotesize\bfseries] at (\durationheaderx, \headerbottom + 0.38) {Adaptation duration (log time)};
  \begin{scope}[shift={(\adaptscopedleft,0)}, xscale=\adaptxscale, shift={(-\adaptlegacyaxisleft,0)}]

  \foreach \tickx in {4.3100,6.4000,8.4900,10.5800} {
    \draw[draw=black!12, line width=0.4pt] (\tickx, \axisy) -- (\tickx, \bandctop - 0.04);
  }

  \adaptbarphase{\rowbottle}{\barh}{\adaptlegacyaxisleft}{6.0749}{orange!70}
  \adaptbarphase{\rowbottle}{\barh}{6.0749}{7.4796}{violet!60}
  \adaptbarphase{\rowbottle}{\barh}{7.4796}{8.4231}{violet!60}
  \adaptbarphase{\rowbottle}{\barh}{8.4231}{8.4766}{black!35}
  \adaptbarphase{\rowbottle}{\barh}{8.4766}{9.1149}{black!35}
  \adaptbarsuccess{\rowbottle}{9.1149}

  \adaptbarphase{\rowbreak}{\barh}{\adaptlegacyaxisleft}{5.7083}{violet!60}
  \adaptbarphase{\rowbreak}{\barh}{5.7083}{6.3731}{orange!70}
  \adaptbarphase{\rowbreak}{\barh}{6.3731}{6.5455}{violet!60}
  \adaptbarphase{\rowbreak}{\barh}{6.5455}{8.2176}{violet!60}
  \adaptbarphase{\rowbreak}{\barh}{8.2176}{9.2507}{black!35}
  \adaptbarphase{\rowbreak}{\barh}{9.2507}{10.2623}{black!35}
  \adaptbarphase{\rowbreak}{\barh}{10.2623}{10.4077}{green!55!black}
  \adaptbarphase{\rowbreak}{\barh}{10.4077}{10.4267}{black!35}
  \adaptbarsuccess{\rowbreak}{10.4267}

  \adaptbarphase{\rowtwotable}{\barh}{\adaptlegacyaxisleft}{4.4089}{cyan!40}
  \adaptbarphase{\rowtwotable}{\barh}{4.4089}{5.3150}{green!55!black}
  \adaptbarphase{\rowtwotable}{\barh}{5.3150}{5.4921}{violet!60}
  \adaptbarphase{\rowtwotable}{\barh}{5.4921}{6.1577}{violet!60}
  \adaptbarphase{\rowtwotable}{\barh}{6.1577}{7.2206}{violet!60}
  \adaptbarphase{\rowtwotable}{\barh}{7.2206}{7.3881}{green!55!black}
  \adaptbarphase{\rowtwotable}{\barh}{7.3881}{7.4679}{violet!60}
  \adaptbarphase{\rowtwotable}{\barh}{7.4679}{7.5365}{orange!70}
  \adaptbarphase{\rowtwotable}{\barh}{7.5365}{8.1480}{blue!45}
  \adaptbarphase{\rowtwotable}{\barh}{8.1480}{8.9049}{green!55!black}
  \adaptbarphase{\rowtwotable}{\barh}{8.9049}{9.0053}{black!35}
  \adaptbarphase{\rowtwotable}{\barh}{9.0053}{9.2552}{violet!60}
  \adaptbarphase{\rowtwotable}{\barh}{9.2552}{10.3363}{violet!60}
  \adaptbarsuccess{\rowtwotable}{10.3363}

  \draw[draw=black!55, line width=0.55pt] (\adaptlegacyaxisleft, \axisy) -- (\adaptlegacyaxisright, \axisy);
  \foreach \tickx/\ticklabel in {4.3100/15 min,6.4000/30 min,8.4900/1 h,10.5800/2 h} {
    \draw[draw=black!55, line width=0.55pt] (\tickx, \axisy - 0.06) -- (\tickx, \axisy + 0.06);
    \node[font=\scriptsize, anchor=north] at (\tickx, \axisy - 0.10) {\ticklabel};
  }
  \node[font=\scriptsize\bfseries, anchor=north] at ({0.5 * (\adaptlegacyaxisleft + \adaptlegacyaxisright)}, \axistitley) {Normalized active adaptation duration (log time)};
  \end{scope}
  \end{scope}

  \inadaptnum{\rowbottle}{9.1149}{1:13:49}
  \inadaptnum{\rowbreak}{10.4267}{1:54:03}
  \inadaptnum{\rowtwotable}{10.3363}{1:50:41}

  \foreach \rowy in {\rowbottle,\rowbreak} {
    \draw[draw=black!12, line width=0.4pt] (\barleft, {\rowy - 0.5 * \rowh}) -- (\figurewidth - 0.10, {\rowy - 0.5 * \rowh});
  }

  \node[bandtitle, fill=yellow!14] at (0.12, \bandctop - 0.06) {This Paper};

  \pgfmathsetmacro{\inlegendy}{\inlegendrowA}
  \inadaptswatch{blue!45, draw=black, line width=0.28pt}{\legAdaptAone}{Setup / structure}
  \inadaptswatch{green!55!black, draw=black, line width=0.28pt}{\legAdaptAtwo}{Approach / locomotion}
  \inadaptswatch{violet!60, draw=black, line width=0.28pt}{\legAdaptAthree}{Opening / manipulation}
  \inadaptswatch{black!35, draw=black, line width=0.28pt}{\legAdaptAfour}{Traversal / completion}
  \inadaptswatch{orange!70, draw=black, line width=0.28pt}{\legAdaptAfive}{Fix / recovery}

  \pgfmathsetmacro{\inlegendy}{\inlegendrowB}
  \node[draw=black, line width=0.28pt, fill=green!55!black, rounded corners=0.6pt, inner xsep=0.8pt, inner ysep=0.4pt, font=\fontsize{6.0}{6.4}\selectfont\bfseries, text=white] at (\legAdaptBone + 0.11, \inlegendy) {$\checkmark$};
  \node[legendtxt, anchor=west] at ({\legAdaptBone + 0.28}, \inlegendy) {First autonomous success};
  \inadaptswatch{cyan!40, draw=black, line width=0.28pt}{\legAdaptBtwo}{Off-robot pre-authoring}
\end{tikzpicture}%
    }
    \par
    }
    \caption{
        In-house real-robot behavior adaptation authoring durations on a log-second axis with authoring phases as colored regions.
        Checkmarks indicate first autonomous success.
    }
    \label{fig:in_house_adaptation_timeline}
\end{figure*}

We next measured sessions where behaviors were composed, adapted, and extended rather than built from scratch.
\autoref{fig:in_house_adaptation_timeline} summarizes adaptation authoring time on a log-second axis with internal milestone structure recovered from screen recordings.

\begin{figure*}[t]
    \centering
    \includegraphics[width=2.0\columnwidth]{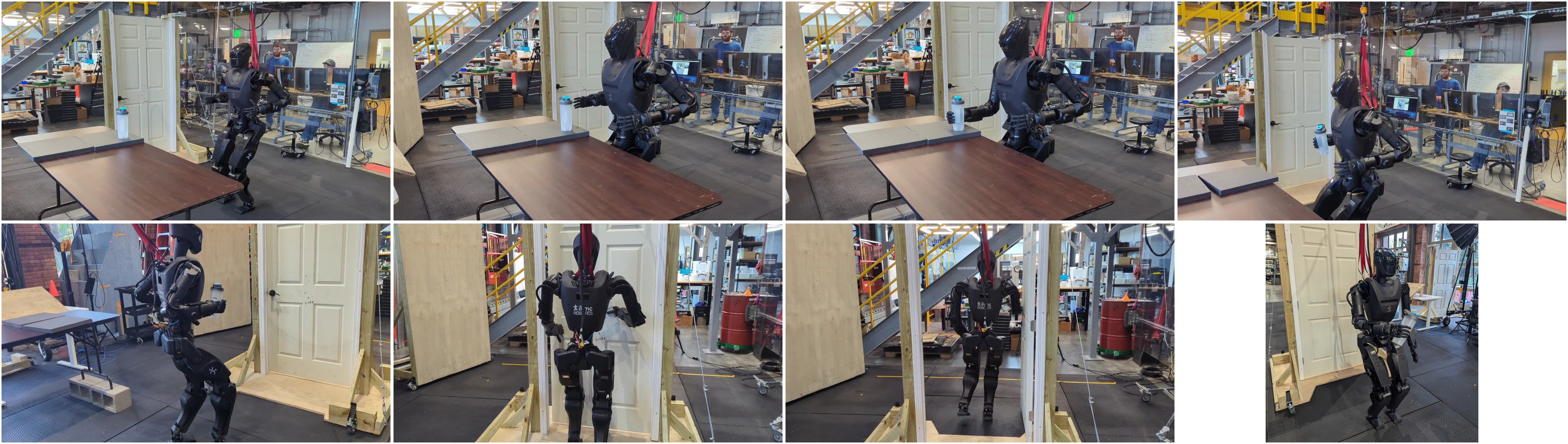}
    \caption{Alex approaching a table, picking up a bottle, carrying it to a door, opening the door with bottle in hand, and walking through the door with it.
    This behavior was the result of a measured authoring session which demonstrates the ability of our system to support quickly composing a new behavior from existing sub-behaviors.}
    \label{fig:h3_bottle_carry_through_authoring}
\end{figure*}

As a composition experiment, we combined an existing bottle pickup behavior with the left push door traversal behavior so Alex would approach a table, pick up the bottle, and carry it through a door in one autonomous run.
Most of the session was centered on wiring the two behaviors together, recovering from a tooling failure while loading the door subtree, and tuning the door opening and traversal routine to keep the bottle held in hand.
We achieved the first autonomous success run at 1 hour, 13 minutes, and 49 seconds of authoring time.
The autonomous run took 64 seconds to grab the bottle from the table and get it through the door, as shown in \autoref{fig:h3_bottle_carry_through_authoring}.

\begin{figure}[h]
    \centering
    \includegraphics[width=0.95\columnwidth]{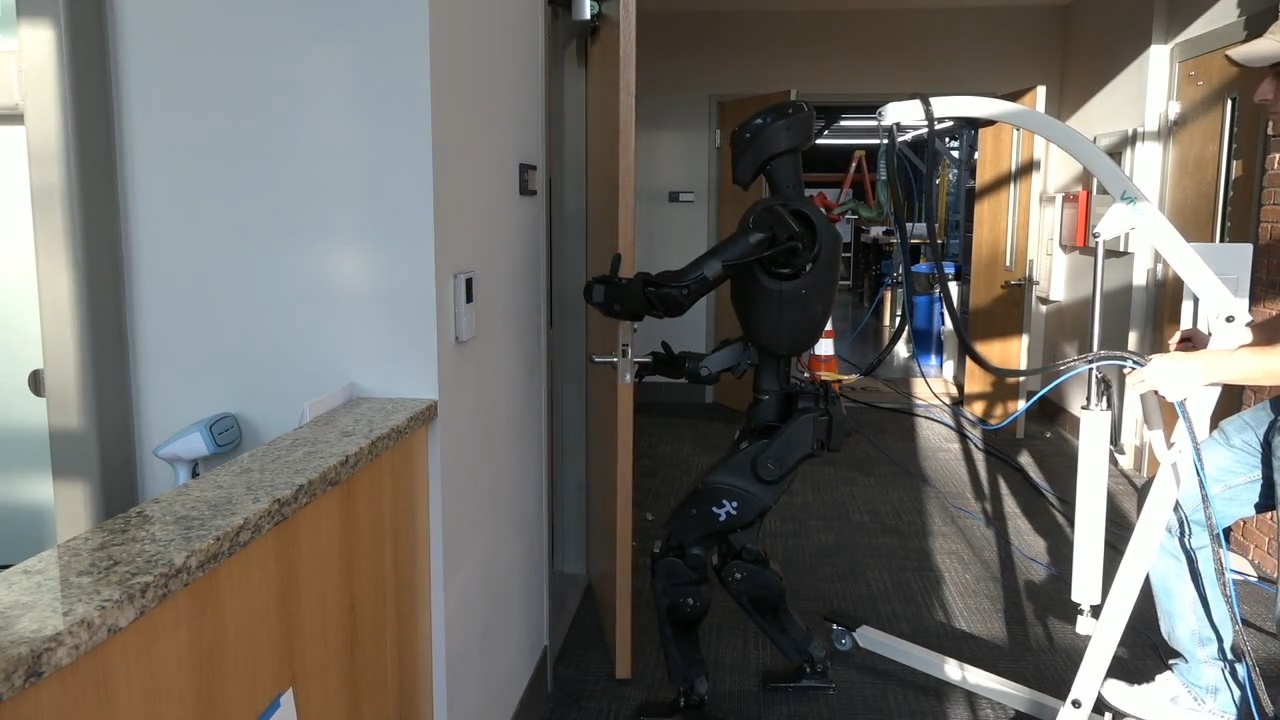}
    \caption{Alex opening the break room left pull door after mirrored adaptation from the lab right pull door behavior.}
    \label{fig:h3_break_room_adaptation}
\end{figure}

We adapted the lab right pull door traversal behavior to the break room left pull door on Alex as shown in \autoref{fig:h3_break_room_adaptation}.
We started by using a mirroring feature in the behavior system which mirrors the definition subtree from right-to-left and vice versa.
However, simply mirroring the primitives was not sufficient because the break room door had a stronger lever spring and the robot's walking quality had degraded.
Most of the 1 hour, 54 minutes, and 3 seconds of measured adaptation time went to tuning the door opening and the traversal footsteps on the real door.
The door opening required more than a screw primitive arm action.
Adding a spine yaw motion after the lever turn was necessary to prevent the door from relatching.
Additionally, we kept having the robot fall during the traversal.
Since the break room door did not have a spring closer, we were able to re-author the footstep plan to a more robust pattern.
This reflects the versatility of our system in being able to quickly work around issues like walking controller tuning which could take much longer to fix.

We extended the reactive single-table ball-sorting behavior from \autoref{sec:resilience} to sort colored balls between two tables with locomotion between stations.
The behavior was meant to put yellow balls in container A on table A and all other colors in container B on table B.
The measured total authoring time was 1 hour, 50 minutes, and 41 seconds, including 15 minutes and 30 seconds of off-robot pre-authoring before the real-robot follow-up session.
The behavior extension showcased the reliability and accuracy of our table approach algorithm, introduced in \autoref{sec:approach_table_object}, as the robot was able to perform three autonomous table approaches in a single run with the centimeter accuracy required to maintain task reachability.
The resulting behavior demonstration autonomously sorted nine of nine balls correctly, with three table approaches and two transitions between tables in 2 minutes 8 seconds.

These measured sessions support hypothesis 3, showing the ability to combine, adapt, and extend existing behaviors to novel tasks in the hours regime.
We credit our adaptation speed to our architectural decisions where we decompose behaviors into reusable primitives, subtrees, and scene actions.
In each case, the work edited part of an existing behavior rather than rebuilding from an empty tree.
The ability to perform these edits and re-test the behavior at runtime speeds up the process even further.

\subsection{Comparative Analysis}
\label{sec:comparative_analysis}

\autoref{fig:in_house_speed_phase_timeline} compares the durations of robot door traversals across this work and selected literature references.
Our door traversals are dramatically faster than the classical model-based results, all of which take more than one minute to traverse a door.
Recent learned systems are also fast: Zhang et al.~\cite{zhang2024learningopentraversedoors} report a 10~s ANYmal door traversal, and DoorMan~\cite{xue2025opening} averages 15.4~s per door.
Our earlier Nadia push-bar door traversal~\cite{calvert2024behavior} cleared the doorway at 14~s; that earlier result, not a measurement from this paper, places the broader runtime-editable lineage in a competitive regime with learned systems.

Platform kinematics help explain some of the speed differences.
Zhang et al.~use a long front-mounted arm on a quadruped with a large yaw workspace, which supports opening doors from a distance and swinging panels open in one motion.

The sensing and compute settings also differ across systems.
Zhang et al.~\cite{zhang2024learningopentraversedoors} use external motion capture and fiducial markers rather than robot vision.
DoorMan~\cite{xue2025opening} runs vision-based neural inference on an off-board computer.
The system evaluated here uses only two color cameras on the robot, with autonomous functionality computed onboard.

Our primary reliability tests were approach-and-opening and loco-manipulation behaviors rather than repeated full doorway traversals.
We did not measure repeated full door traversals because the walk-through component was not reliable enough during the available test window without risking hardware damage.
The approach-and-opening trials still indicate that the behavior architecture, perception, and manipulation control can achieve high reliability; remaining traversal unreliability was dominated by walking controller performance rather than the architecture described in this paper.

Most of the literature reports reliability in a similar envelope.
Jain and Kemp~\cite{Jain2008dooropening,Jain2010EPC} report 21/24 and 37/40 success-trial counts, Chitta et al.~\cite{2010ChittaDoorOpening} report 5/5 for push and pull types, Schulze et al.~\cite{Schulze_2025} report 109/115 and 37/42, and Zhang et al.~\cite{zhang2024learningopentraversedoors} report 20/20 pull and 18/20 push.
Although we reported no failures in the approach-and-opening trials above, that does not imply the system is failure-free in general; the same caveat applies to short repeated-run reports throughout the literature.
DoorMan~\cite{xue2025opening} reports an 83\% reliability figure; the publication does not include the corresponding attempt and trial counts, so we do not attempt a direct numerical comparison.

\begin{figure*}[t]
\centering
\resizebox{\textwidth}{!}{
\newcommand{\procsteprow}[5]{%
  \begin{tabular}{@{}>{\raggedright\arraybackslash}p{\proctextw}@{}>{\raggedleft\arraybackslash}p{\proctimew}@{}}%
    \textbf{\textcolor{#5}{#1.}} #2 & \textcolor{#4}{\fontsize{7.4}{8.0}\selectfont\bfseries #3}%
  \end{tabular}%
}

\newcommand{\proclooparrow}[2]{%
  \draw[procloop]
    (#1.north east)
    .. controls ($(#1.north east) + (0.58, 0.06)$) and ($(#2.east) + (0.58, 0.06)$)
    .. (#2.east);%
}
\newcommand{\procloopself}[1]{%
  \draw[procloop]
    (#1.east)
    .. controls ($(#1.east) + (0.62, 0.20)$) and ($(#1.north east) + (0.62, 0.20)$)
    .. (#1.north east);%
}

\begin{tikzpicture}[
    x=1cm,
    y=1cm,
    every node/.style={outer sep=0pt},
    procboxthesis/.style={
      draw=blue!55!black,
      fill=blue!14,
      rounded corners=3pt,
      align=left,
      inner xsep=4pt,
      inner ysep=4pt,
      minimum height=0.58cm,
      font=\fontsize{6.9}{7.8}\selectfont,
      text=black!90
    },
    procboxdoorman/.style={
      draw=orange!75!black,
      fill=white!92,
      rounded corners=3pt,
      align=left,
      inner xsep=4pt,
      inner ysep=4pt,
      minimum height=0.58cm,
      font=\fontsize{6.9}{7.8}\selectfont,
      text=black!90
    },
    procarr/.style={->, >=stealth, line width=0.65pt, draw=blue!50!black},
    procarrdm/.style={->, >=stealth, line width=0.65pt, draw=orange!70!black},
    procloop/.style={
      ->,
      >=stealth,
      dashed,
      line width=0.65pt,
      draw=violet!75!black,
      rounded corners=3pt
    },
    looplbl/.style={font=\fontsize{5.6}{6.2}\selectfont\itshape, text=violet!75!black},
    looplegendlbl/.style={font=\fontsize{7.0}{7.8}\selectfont\itshape, text=violet!75!black},
    looplegendline/.style={
      procloop,
      line width=0.85pt
    },
    paneltitle/.style={
      font=\footnotesize\bfseries,
      anchor=west,
      text=blue!55!black
    },
    colhdrthesis/.style={
      font=\scriptsize\bfseries,
      anchor=south,
      align=center,
      text=blue!60!black
    },
    colhdrdm/.style={
      font=\scriptsize\bfseries,
      anchor=south,
      align=center,
      text=orange!80!black
    },
    colsub/.style={
      font=\fontsize{6.4}{7.0}\selectfont,
      anchor=north,
      align=center,
      text=black!55
    },
    panelresult/.style={
      font=\fontsize{7.8}{8.6}\selectfont\bfseries,
      anchor=center,
      align=center
    }
  ]

  \def\figw{14.55}
  \def\panelAh{7.49}
  \def\colgap{0.44}
  \def\boxw{6.55}
  \def\colw{6.88}
  \def\leftcx{0.14 + 0.5 * \colw}
  \def\rightcx{0.14 + \colw + \colgap + 0.5 * \colw}
  \def\proctextw{3.50cm}
  \def\proctimew{2.90cm}
  \def\stepgap{0.26}
  \def\headerbandh{1.30}
  \def\paneltitlepad{0.30}
  \def\colhdry{1.06}
  \def\colsubsep{-3.5pt}
  \def\footertotalsep{0.44}
  \def\panelbottomy{1.32}
  \def\looplegendpad{0.46}

  \def\coltimeblue{blue!65!black}
  \def\coltimeviolet{violet!75!black}
  \def\coltimeorange{orange!85!black}
  \def\coltimegreen{green!50!black}
  \def\colnumthesis{blue!65!black}
  \def\colnumdoorman{orange!80!black}

  \pgfmathsetmacro{\panelAtop}{\panelbottomy + \panelAh}
  \pgfmathsetmacro{\panelAbottom}{\panelbottomy}
  \pgfmathsetmacro{\figheight}{\panelAtop + \looplegendpad + 0.08}

  \pgfmathsetmacro{\flowAtop}{\panelAtop - \headerbandh}
  \pgfmathsetmacro{\splitx}{0.5 * \figw}
  \pgfmathsetmacro{\lefthalfcx}{0.5 * \splitx}
  \pgfmathsetmacro{\righthalfcx}{0.5 * (\splitx + \figw)}

  \path[draw=blue!35, fill=cyan!12, rounded corners=5pt, line width=0.85pt]
    (0, \panelAbottom) rectangle (\figw, \panelAtop);

  \path[fill=orange!7, opacity=0.88]
    (\splitx, \panelAbottom) rectangle (\figw, \panelAtop);

  \node[paneltitle] at (0.16, \panelAtop - \paneltitlepad)
    {Creation from Scratch};
  \node[colhdrthesis] (aHdrTh) at (\leftcx, \panelAtop - \colhdry) {This Thesis, Scratch Left Push Door, 2026-02-22};
  \node[colsub, below=\colsubsep of aHdrTh] {on-robot edit-test, measured};
  \node[colhdrdm] (aHdrDm) at (\rightcx, \panelAtop - \colhdry) {DoorMan~\cite{xue2025opening}};
  \node[colsub, below=\colsubsep of aHdrDm] {sim-to-train pipeline, off-board, estimated};

  \node[procboxthesis, anchor=north, text width=\boxw cm] (aL1) at (\leftcx, \flowAtop - 0.08)
    {\procsteprow{1}{Create empty behavior sequence}{0:00:00}{\coltimeblue}{\colnumthesis}};
  \node[procboxthesis, anchor=north, text width=\boxw cm] (aL2) at ($(aL1.south) + (0,-\stepgap)$)
    {\procsteprow{2}{Add approach + door-panel scene action}{0:14:33--0:24:22}{\coltimeblue}{\colnumthesis}};
  \node[procboxthesis, anchor=north, text width=\boxw cm] (aL3) at ($(aL2.south) + (0,-\stepgap)$)
    {\procsteprow{3}{Author handle turn and opening}{1:21:14--1:24:37}{\coltimeblue}{\colnumthesis}};
  \node[procboxthesis, anchor=north, text width=\boxw cm] (aL4) at ($(aL3.south) + (0,-\stepgap)$)
    {\procsteprow{4}{Add traversal footsteps + arm push}{1:39:21--1:49:24}{\coltimeblue}{\colnumthesis}};
  \node[procboxthesis, anchor=north, text width=\boxw cm] (aL5) at ($(aL4.south) + (0,-\stepgap)$)
    {\procsteprow{5}{Edit-test fixes + stop condition}{1:50:11--1:58:15}{\coltimeviolet}{\colnumthesis}};
  \node[procboxthesis, anchor=north, text width=\boxw cm] (aL6) at ($(aL5.south) + (0,-\stepgap)$)
    {\procsteprow{6}{First fully automatic success}{1:59:48}{\coltimegreen}{\colnumthesis}};

  \foreach \i/\j in {aL1/aL2, aL2/aL3, aL3/aL4, aL4/aL5, aL5/aL6} {
    \draw[procarr] (\i.south) -- (\j.north);
  }
  \proclooparrow{aL5}{aL4}

  \node[procboxdoorman, anchor=north, text width=\boxw cm] (aR1) at (\rightcx, \flowAtop - 0.08)
    {\procsteprow{1}{Build randomized physics simulation}{est.\ days}{\coltimeorange}{\colnumdoorman}};
  \node[procboxdoorman, anchor=north, text width=\boxw cm] (aR2) at ($(aR1.south) + (0,-\stepgap)$)
    {\procsteprow{2}{Tune PPO teacher (privileged state)}{est.\ many h}{\coltimeorange}{\colnumdoorman}};
  \node[procboxdoorman, anchor=north, text width=\boxw cm] (aR3) at ($(aR2.south) + (0,-\stepgap)$)
    {\procsteprow{3}{DAgger distillation to RGB student}{est.\ many h}{\coltimeorange}{\colnumdoorman}};
  \node[procboxdoorman, anchor=north, text width=\boxw cm] (aR4) at ($(aR3.south) + (0,-\stepgap)$)
    {\procsteprow{4}{GRPO fine-tuning}{est.\ many h~+~overnight}{\coltimeorange}{\colnumdoorman}};
  \node[procboxdoorman, anchor=north, text width=\boxw cm] (aR5) at ($(aR4.south) + (0,-\stepgap)$)
    {\procsteprow{5}{Real-robot hardware validation}{est.\ $\sim$1 h}{\coltimeorange}{\colnumdoorman}};

  \foreach \i/\j in {aR1/aR2, aR2/aR3, aR3/aR4, aR4/aR5} {
    \draw[procarrdm] (\i.south) -- (\j.north);
  }
  \procloopself{aR2}

  \node[panelresult, text=blue!60!black] (totTh) at ($(aL6.south) + (0,-\footertotalsep)$)
    {Total: \textbf{2 hours}};
  \node[panelresult, text=orange!85!black] at (aR5 |- totTh)
    {Total: est.\ \textbf{2--6 days + 1 overnight + 1 h}};

  \draw[draw=black!38, line width=0.55pt, densely dotted]
    (\splitx, \panelAbottom) -- (\splitx, \panelAtop);

  \pgfmathsetmacro{\looplegendy}{\panelAbottom - 0.30}
  \draw[looplegendline] (0.16, \looplegendy) -- ++(0.52, 0);
  \node[looplegendlbl, anchor=west] at (0.74, \looplegendy) {iterative edit / train retry};

\end{tikzpicture}}
\caption{Authoring a door behavior from scratch in our system (measured) versus DoorMan (estimated).}
\label{fig:hero_authoring_process_comparison}
\end{figure*}
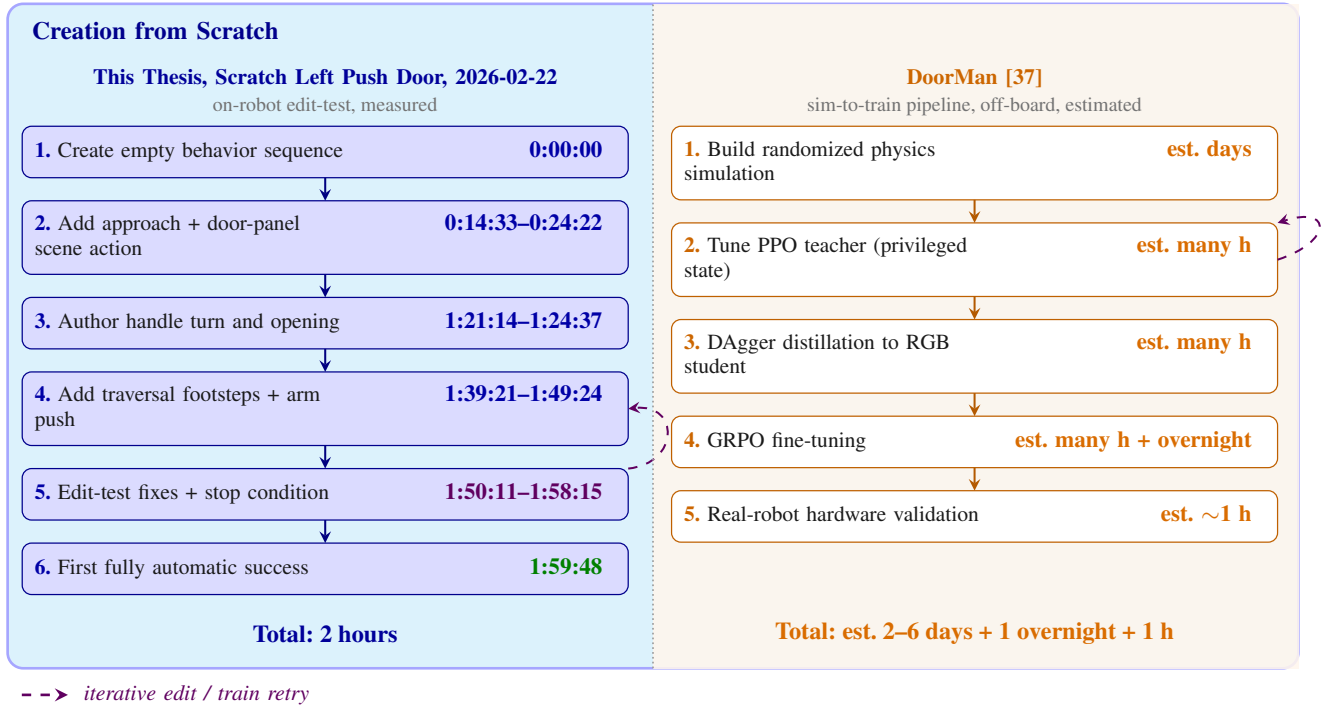

This work is unique in reporting measured behavior authoring duration for loco-manipulation.
We report both creation from scratch and adaptation of existing behaviors.
We did not find documented authoring-duration results for comparable loco-manipulation behaviors in the literature.

As an illustrative comparison, \autoref{fig:hero_authoring_process_comparison} contrasts a measured push-door creation from scratch in our system (6 steps in 2 hours) with an estimated DoorMan pipeline based on the steps presented in that paper.
Under that estimate, the runtime-editable workflow is substantially faster for creating and adapting behaviors, because edits can be localized and retested online.
By contrast, adapting a DoorMan-style policy typically requires repeating simulation setup, policy tuning, and overnight retraining.
The difference follows from decomposing behaviors into reusable primitives, subtrees, and scene actions: changing a pull door to a sliding door may require a few adjusted actions and perhaps an alternate YOLO model for the handle, rather than a full policy refit.

\section{Discussion and Conclusion}
\label{ch:pontification}

A model-based behavior stack combining Coactive Design principles for humanoid operation, affordance templates, a tree structure for organization and logic, and a behavior-managed perception scene has proved effective for the loco-manipulation tasks evaluated here.
Over a decade of humanoid autonomy work in the broader community, systems have moved from large teleoperation teams for multi-task demos toward smaller teams bringing up autonomous loco-manipulation behaviors in hours; technology progress across the full stack contributes heavily, and we argue that runtime-editable behavior architecture is one enabling factor.

Some in the field have now abandoned similar approaches in pursuit of learning-based ones.
However, we think a lot of tasks can be solved both ways.
The two can also be combined, because our system supports learning-based low-level routines as nodes of the tree.
In fact, with the recently increased capabilities of intelligent coding assistance, the more classical, model-based approaches may even be able to keep pace with learned systems.

\subsection{Strong Points}

We think the most impactful design elements of our system are robot-locality and runtime-editability.
Running the compute locally robustifies the system by decoupling it from external digital disturbances such as communications failures and compute failures on remote servers.
Editing behaviors online has enabled rapid bringup of new behaviors and modification, composition, and extension of existing behaviors, as we have shown.

We also think the interface between the behavior system and the robot controller works well.
The ability to schedule asynchronous footstep, arm, leg, spine, and neck commands enables a high level of behavioral expressiveness.
It also supports the integration of classical planners and AI-assisted behavior composition.

Behavior-time perception through authorable scene actions has also been crucial.
Robot perception systems have a lot of important and non-intuitive quality levels.
Furthermore, these quality levels can vary with lighting, situation, and hardware differences.
The metrics of quality include depth accuracy, semantic object detection confidence, latency, and frequency.
Our design designates an expert human operator to evaluate and exploit available quality levels for each behavior situation to achieve the goal at hand.

\subsection{Weak Points}

There are important caveats, however, with our current system.
For one, our object-on-table manipulation demos center around a bottle pickup and balls.
We do not yet have good performance in directly estimating the orientation of objects.
This is why our premier sorting demos use spherical balls.
Since they are symmetrical objects, only position is required to grasp them from any direction.
We have preliminary results using FoundationPose for object orientation estimation, but it is not yet working well enough for repeatable tests.

Another weak point is the lack of support for executing sequence subtrees that return when they are done like subroutines.
This would solve two existing problems in the system.
For one, it would solve the problem of having two of the same JSON file loaded in the tree at once.
When the same JSON file is loaded twice in the tree, changes to one instance are not reflected in the other, and saving one can overwrite the other.
If sequences could be run as a subtree and return when done, identical subtrees throughout the system could be instantiated once instead of duplicated.
This functionality would also add the ability to call subroutines from multiple places.

\subsection{Natural Next Steps}

Throughout this work, we have wanted to add force-based action primitives.
The codebase even includes a ``wrench action'' node, which was used to pick up boxes.
Force-based action definitions could provide proprioceptive capabilities such as detecting when the door handle is all the way turned.
It could also enable compliance when manipulating articulated objects.
However, we still have not had the time to approach it properly from a controls perspective.

Another area of work that would be nice to explore is grasp planning.
For example, when operating tools designed for humans, it would be helpful to plan a 5-finger grasp for them instead of having to manually specify finger configurations.
A grasp planner could be built into our Ability Hand action node to automatically generate finger positions based on behavioral context.
The behavior author could also use their own hands to demonstrate an example grasp technique to seed the planner.

Taking a step back and looking at our system as a whole, there is also an important gap in handling open world environments.
We focus on a human operator's ability to author robot behaviors rather than robot-generated ones.
The lack of a generative ability on the robot, where it could ``author'' its own behavior, limits our system to well-thought-out, human-constructed scenarios.
Large vision and language models are becoming increasingly capable and may be able to fill in for or assist the operator in some cases.

\section*{Acknowledgements}
We would like to thank Jerry Pratt, Stephen Hart, Brent Venable, William Howell, Jordan Accardo, Thomas McKenna, and the IHMC Robotics team, without whom this work would not have been possible.

\section*{Source Code and Media}
Our implementation can be found on GitHub at \url{https://github.com/ihmcrobotics}.
The accompanying video can be found at \url{https://www.youtube.com/playlist?list=PLJK5CTyotYqsfgfnXb-09YNFeBose6uEY}.

\bibliography{bibliography}

\end{document}